\pdfoutput=1
\documentclass[10pt, a4paper]{article}

\usepackage{lrec-coling2024} % this is the new style

\usepackage{times}
\usepackage{latexsym}

\usepackage[T1]{fontenc}
\usepackage[utf8]{inputenc}

\usepackage{microtype}

\usepackage{inconsolata}

\usepackage{amsmath}
\usepackage{graphicx}
\usepackage{float}
\usepackage[export]{adjustbox}
\usepackage{subcaption}
\usepackage{booktabs}
\usepackage{hyperref}
\usepackage{parskip}
\usepackage{longtable}
\usepackage{tikz}
\usetikzlibrary{positioning}
\usetikzlibrary{calc}
\usetikzlibrary{fit}
\usetikzlibrary{shapes.multipart}
\tikzset{
    *|/.style={
        to path={
            (perpendicular cs: horizontal line through={(\tikztostart)},
                                 vertical line through={(\tikztotarget)})
            -- (\tikztotarget) \tikztonodes
        }
    }
}

\title{Instructions for *ACL Proceedings}

\author{Sam Spilsbury \\
  Department of Computer Science \\
  Aalto University \\
  Espoo, Finland \\
  \texttt{sam.spilsbury@aalto.fi} \\\And
  Pekka Martinen \\
  Department of Computer Science \\
  Aalto University \\
  Espoo, Finland \\
  \texttt{pekka.martinen@aalto.fi} \\\And
  Alexander Ilin \\
  Department of Computer Science \\
  Aalto University \\
  Espoo, Finland \\
  \texttt{alexander.ilin@aalto.fi} \\}

\maketitle
\begin{abstract}
In-Context-learning and few-shot prompting are viable methods to induce certain types of compositional behaviour. However, these methods can be very sensitive to the choice of support examples used. Choosing good supports from the training data for a given test query is already a difficult problem, but in some cases solving this may not even be enough. We consider a grounded language learning problem (gSCAN) where it is difficult to search for helpful supports in some cases. We design an agent which instead generates possible supports which are relevant to the test query and current state of the world, then uses these supports via in-context-learning to solve the test query. We show substantially improved performance  on a previously unsolved compositional behaviour split without a loss of performance on other splits. {\color{black}We also show that our approach scales to a more challenging version of the dataset with natural language instructions.}
\end{abstract}

\title{Generating Demonstrations for In-Context Compositional Generalization in Grounded Language Learning}

\begin{document}
\section{Introduction}

We want autonomous agents to have the same compositional understanding of language that humans do \citep{book/chomsky/1957, conf/atal/Tenenbaum18}.
Without this understanding, the sample complexity required to train them for a wide range of compositions of
instructions would be very high \citep{conf/icml/Sodhani0P21, conf/corl/JangIKKELLF21}. Naturally, such compositional generalization has received interest from both
the language and reinforcement learning communities. ``Compositional Generalization" can be divided into several
different sub-skills, for example being able to reason about object properties compositionally \citep{conf/aaai/ChaplotSPRS18, conf/emnlp/QiuH0SS21}, composing sub-instructions
into a sequence \citep{conf/naacl/LogeswaranFLL22, conf/iclr/Min22} or generating novel outputs according to novel inputs made up of familiar components \citep{conf/icml/LakeB18}.

A long line of work and many different datasets show that Deep Learning approaches do not always achieve
such compositional generalization, especially in the case of novel output sequences.
Some solutions to make up for this deficiency include modular architectures,
data augmentation, and sparsity. A recent line of work concerns in-context learning (ICL). Instead of just providing
a query and asking for the target directly, a few examples of query-target pairs, called \textit{supports}, are provided
along with the query. In the compositional generalization case, we cannot provide out-of-distribution examples
showing the expected behaviour exactly, but as long as the examples are \textit{relevant} in that they
cover the correct elements of the problem space, then compositional generalization is possible. This immediately begs
the follow up question of how such relevant examples should be generated for each query. Most of the prior work in this
area takes one of four approaches: searching for near-neighbours to the query input \cite{conf/emnlp/PasupatZG21}; searching for solutions to subproblems (assuming that the subproblems are known) \cite{conf/naacl/Yang22}, searching for near-neighbours of the initial predicted output \cite{conf/coling/Zemlyanskiy22} and chain-of-thought prompting \cite{conf/neurips/Wei22}.

\begin{figure*}[ht]
\resizebox{\textwidth}{!}{
\begin{tikzpicture}[
    title/.style={font=\fontsize{6}{6}\color{black!50}\ttfamily},
    node distance = 10mm, 
]
    \node [fill=black!10,rounded corners, inner sep=3pt] (query) {
        \begin{tikzpicture}[node distance=0mm]
            \node [anchor=west] (title) {\footnotesize{Query}};
            \node [anchor=north, below=of title.south] (state)
        {\includegraphics[width=.35\textwidth,keepaspectratio=1,trim={4cm 1cm, 3.5cm 1cm},clip]{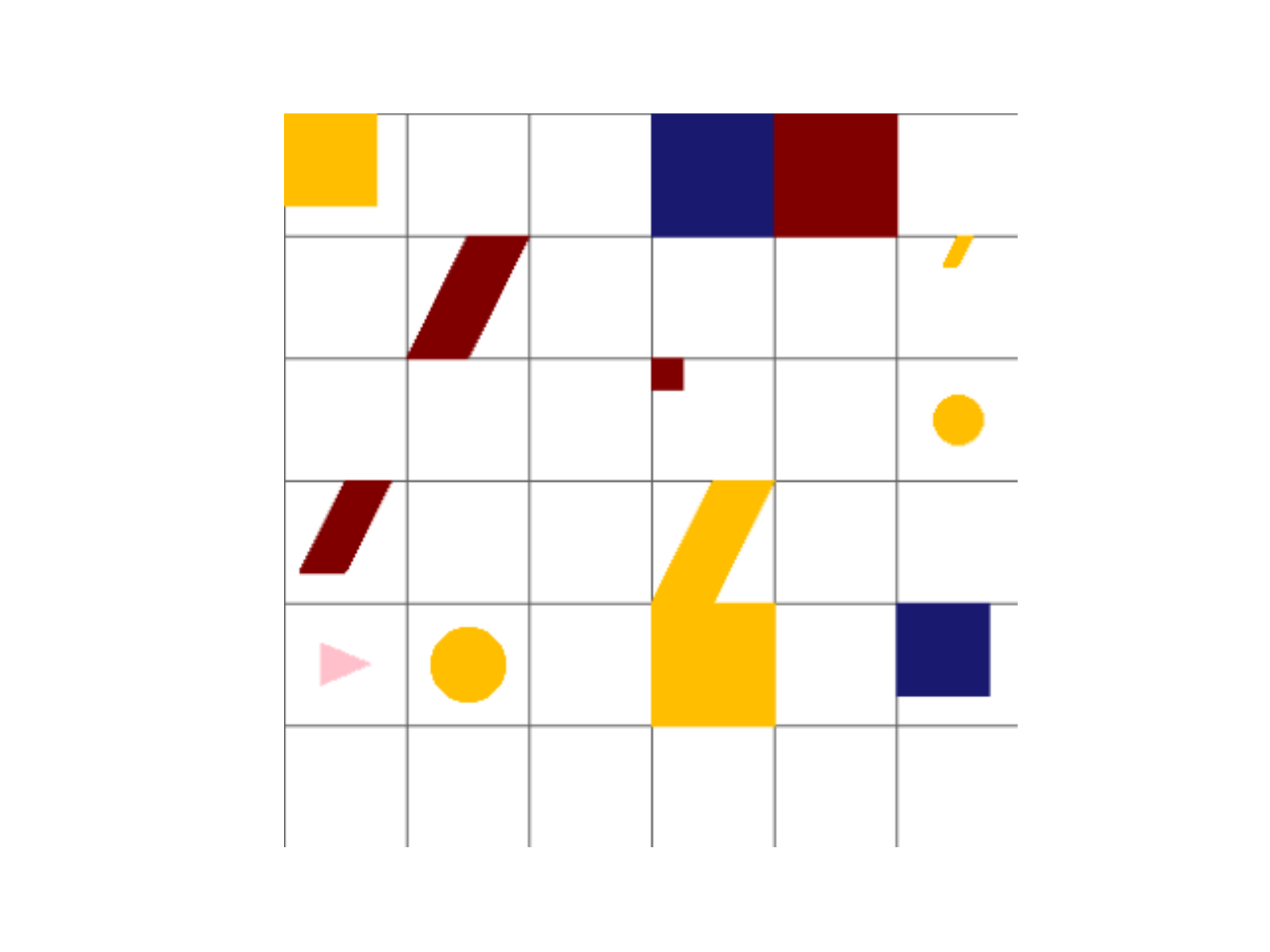}};
            \node[inner sep=0pt, below=of state.south] (query)
        {\footnotesize{$I^Q$ = ``spin and pull a small yellow cylinder"}};
        \end{tikzpicture}
    };
    \node [fill=black!10, rounded corners, inner sep = 5pt, right=of query.east, minimum width=50mm, anchor=south, rotate=90] (ig) {Instruction Generator};
    \node [fill=green!10, rounded corners, inner sep = 5pt, right= 7mm of ig.east, anchor=west, yshift=7mm] (i1) {\footnotesize{$I_1$ = carefully zigzag and pull a small yellow cylinder (0.46)}};
\node [fill=green!10, rounded corners, inner sep = 5pt, below=7mm of i1.west, anchor=west] (i2) {\footnotesize{$I_3$ = spin and push a small yellow cylinder (0.46) }};
\node [fill=yellow!10, rounded corners, inner sep = 5pt, below=7mm of i2.west, anchor=west] (i3) {\footnotesize{$I_5$ = take a cautious zigzagging path to a small yellow cylinder (0.35) }};
\node [fill=green!10, rounded corners, inner sep = 5pt, below=7mm of i3.west, anchor=west] (i4) {\footnotesize{$I_6$ = carefully spin and push a small yellow cylinder (0.33) }};
\node [fill=green!10, rounded corners, inner sep = 5pt, below=7mm of i4.west, anchor=west] (i5) {\footnotesize{$I_8$ = spin and nudge a small yellow cylinder (0.29) }};
\node [fill=green!10, rounded corners, inner sep = 5pt, below=7mm of i5.west, anchor=west] (i6) {\footnotesize{$I_{13}$ = spin and pull a big yellow cylinder (0.19) }};
\node [fill=green!10, rounded corners, inner sep = 5pt, below=7mm of i6.west, anchor=west] (i7) {\footnotesize{$I_{16}$ = gently pull a small yellow cylinder (0.19) }};
\node [fill=red!10, rounded corners, inner sep = 5pt, below=7mm of i7.west, anchor=west] (i8) {\footnotesize{$I_{18}$ = spin and carefully pull a small green cylinder (0.18) }};
\node [fill=red!10, rounded corners, inner sep = 5pt, below=7mm of i8.west, anchor=west] (i9) {\footnotesize{$I_{21}$ = spin and carefully pull a small red cylinder (0.16) }};
\node [fill=red!10, rounded corners, inner sep = 5pt, below=7mm of i9.west, anchor=west] (i10) {\footnotesize{$I_{22}$ = spin and carefully pull a small blue cylinder (0.15) }};
    \node [fill=black!10, rounded corners, inner sep = 5pt, right=11cm of query.east, minimum width=50mm, anchor=south, rotate=90] (at) {Transformer};
    \node [fill=green!10, rounded corners, inner sep = 5pt, right= 7mm of at.east, anchor=west, yshift=7mm] (a1) {\footnotesize{$A_1$ = (WALK LTURN WALK RTURN)(3)WALK(2) PULL(3)}};
\node [fill=green!10, rounded corners, inner sep=5pt, below=7mm of a1.west, anchor=west] (a2) {\footnotesize{$A_3$ = LTURN(4) (WALK LTURN(4))(5) LTURN (WALK LTURN(4))(3) PUSH}};
\node [fill=green!10, rounded corners, inner sep=5pt, below=7mm of a2.west, anchor=west] (a3) {\footnotesize{$A_5$ = (WALK LTURN WALK RTURN)(3) WALK(2)}};
\node [fill=green!10, rounded corners, inner sep=5pt, below=7mm of a3.west, anchor=west] (a4) {\footnotesize{$A_6$ = LTURN(4) (WALK LTURN(4))(4) LTURN (WALK LTURN(4))(3) PUSH}};
\node [fill=green!10, rounded corners, inner sep=5pt, below=7mm of a4.west, anchor=west] (a5) {\footnotesize{$A_8$ = LTURN(4) (WALK LTURN(4))(5) LTURN (WALK LTURN(4))(3) PUSH}};
\node [fill=black!10, rounded corners, inner sep=5pt, below=7mm of a5.west, anchor=west] (a6) {\footnotesize{$A_{13}$ = LTURN(4) (WALK LTURN(4))(3) LTURN WALK}};
\node [fill=black!10, rounded corners, inner sep=5pt, below=7mm of a6.west, anchor=west] (a7) {\footnotesize{$A_{16}$ = (WALK STAY)(4) LTURN (WALK STAY)(3)}};
\node [fill=black!10, rounded corners, inner sep=5pt, below=7mm of a7.west, anchor=west] (a8) {\footnotesize{$A_{18}$ = LTURN(4) (WALK LTURN(4))(5) LTURN (WALK LTURN(4))(3) WALK PULL LTURN(3) PUSH}};
\node [fill=black!10, rounded corners, inner sep=5pt, below=7mm of a8.west, anchor=west] (a9) {\footnotesize{$A_{21}$ = LTURN(5) WALK PULL LTURN(3) PUSH}};
\node [fill=black!10, rounded corners, inner sep=5pt, below=7mm of a9.west, anchor=west] (a10) {\footnotesize{$A_{22}$ = LTURN(4) (WALK LTURN(4))(5) LTURN (WALK LTURN(4))(3) WALK PULL LTURN(3) PUSH}};
    \end{tikzpicture}
}
\caption{Generating demonstrations for use with for gSCAN with DemoGen. The Instruction Generator takes as input the current state and $I_q$ and produces similar instructions $I_1, ... I_{k}$ likely to occur in the same state, sorted by likelihood (parens). A Transformer trained on the training data generates the corresponding actions in that state. Some instructions are more helpful than others. Instructions in {green, $I_{1, 3, 6, 8, 13, 16}$} show both the correct object in $I_q$ and also either one of the verb or adverb. Instructions in {yellow, $I_5$} show the correct object, an irrelevant verb and adverb combination. Instructions in {red, $I_{18, 21, 22}$} show a different object to the target one. Actions in grey $A_{13, 16, 18, 21, 22}$ show an incorrect target sequence. As long as the instructions and actions in {green} are included in the support set, a sufficiently powerful model can use them and ignore the other supports. Duplicates omitted.}
\label{fig:oracle}
\end{figure*}

We suggest that in the Grounded Language Learning case, these approaches might not be sufficient to make compositional generalization
by ICL work. In Grounded Language Learning, the outputs are conditional not only on the \textit{query}, but also on
the \textit{state} of the world. Searching for nearby examples in the input space thus becomes problematic. Using the query alone
means that it is unlikely that \textit{state-relevant} examples will be retrieved. The complexity of the state
space is so large that there might not even be other examples in the same state and finding \textit{similar} states is challenging
because small changes in the state can result in large changes to the target sequence. For example, a change to the position of the target
object in an object reaching task, where all other objects stay in the same position, results in a large change to the target sequence,
but a large change in the position of other objects results in little-to-no change. Searching for nearby examples in the \textit{output space} like \cite{conf/coling/Zemlyanskiy22}
is more promising, but it relies on the assumption that you can find outputs in the training data which happen to match what the state requires. We show in this work that on a well-known Grounded Language Learning
benchmark (gSCAN), it is difficult to make a retrieval-based strategy that works well in all cases.

We suggest another way to approach the problem, which is to generate the supports. We call our method DemoGen. It first generates near neighbours of the query
as support inputs, ranks them by their applicability to the current state, then \textit{generates} the corresponding support outputs conditioned on the
current state  (Figure \ref{fig:oracle}). The generated supports are used for Transformer ICL at test time (Figure \ref{fig:transformer}). The generation and ranking processes are trained using access only to the training data. The supports inputs and outputs generated by our method are typically congruent with the underlying environment rules.
It is possible to generate an out of distribution support input, or a support that might not be relevant to the query at hand, or even a support with an incorrect demonstration, but
we show that in practice, this does not matter all that much as long all the relevant supports are generated. Through our experiments, we show that
our method is able to unlock better compositional generalization performance on a challenging split of gSCAN, without sacrificing significant amounts of performance
in other cases. {\color{black}Furthermore, we show that our approach can also scale to
more challenging problems with instructions that resemble natural language.}

\section{Related work}
\label{sec:related_work}

\subsection{In-context and Meta-learning for Compositional Generalization}

\begin{figure*}[t]
\centering
\begin{tikzpicture}[scale=1.0,transform shape,
  input/.style = {draw, fill=blue!20, rounded corners, minimum height=7mm},
  output/.style = {draw, fill=green!20, rounded corners},
  special/.style = {draw, fill=red!20, rounded corners},
  embedding/.style = {draw, fill=yellow!20, rounded corners},
  transformer/.style = {draw, fill=gray!20, rounded corners}
  ]

  \node[transformer, minimum width=85mm] (encoder) at (0, 0) {Encoder};

  \node[input, below=1.8cm of encoder.west, xshift=0.4cm] (s1) {$S_1$};
  \node[input, right=0.3cm of s1] (i1) {$I_1$};
  \node[input, right=0.3cm of i1] (a1) {$A_1$};
  \node[special, above=0.5cm of a1] (a1p) {$P$};
  \node[right=0.3cm of a1] (edots) {$...$};
  \node[input, right=0.3cm of edots] (sn) {$S_n$};
  \node[input, right=0.3cm of sn] (in) {$I_n$};
  \node[input, right=0.3cm of in] (an) {$A_n$};
  \node[special, above=0.5cm of an] (anp) {$P$};
  \node[input, right=0.3cm of an] (sq) {$S_q$};
  \node[input, right=0.3cm of sq] (iq) {$I_q$};
  
  \draw[->] (s1.north) -- ([yshift=1.4cm] s1.north);
  \draw[->] (i1.north) -- ([yshift=1.4cm] i1.north);
  \draw[->] (a1.north) -- ([yshift=0.4cm] a1.north);
  \draw[->] (a1p.north) --([yshift=0.35cm] a1p.north);
  \draw[->] (sn.north) -- ([yshift=1.4cm] sn.north);
  \draw[->] (in.north) -- ([yshift=1.4cm] in.north);
  \draw[->] (an.north) -- ([yshift=0.4cm] an.north);
  \draw[->] (anp.north) --([yshift=0.35cm] anp.north);
  \draw[->] (sq.north) -- ([yshift=1.4cm] sq.north);
  \draw[->] (iq.north) -- ([yshift=1.4cm] iq.north);

  \node[transformer, minimum width=40mm, right=0.3 cm of encoder] (decoder) {Decoder};
  \node[special, below=0.5cm of decoder.south, minimum width=40mm] (decoderp) {$P$};
  \node[input, below=1.8cm of decoder.west, xshift=0.5cm] (sos) {[sos]};
  \node[input, right=0.3cm of sos] (t1) {$a^1_q$};
  \node[right=0.3cm of t1] (tdots) {$...$};
  \node[input, right=0.3cm of tdots] (tn) {$a^n_q$};

  \draw[->] (sos.north) -- ([yshift=0.4cm] sos.north);
  \draw[->] (t1.north) -- ([yshift=0.4cm] t1.north);
  \draw[->] (tn) -- ([yshift=0.4cm] tn.north);

  \draw[->] ([yshift=1.0cm] sos.north) -- ([yshift=1.4cm] sos.north);
  \draw[->] ([yshift=1.0cm] t1.north) -- ([yshift=1.4cm] t1.north);
  \draw[->] ([yshift=1.0cm] tn.north) -- ([yshift=1.4cm] tn.north);
  
\end{tikzpicture}
\caption{The model architecture for sequence-to-sequence ICL. Each support state $S_1, ..., S_n$, support instruction $I_1, ..., I_n$ and corresponding support targets $A_1, ..., A_n$, as well as the query state $S_q$ and query instruction $I_q$ are used as inputs to a Transformer Encoder (along with positional encoding). Right-shifted query targets $a_q^1, ..., a_q^n$ are used as inputs to a Transformer Decoder. Both the support targets and query targets use the same random permutation on every training step.}
\label{fig:transformer}
\end{figure*}

Meta-learning and ICL are promising approaches for compositional generalization
in sequence generation tasks. In this paradigm, a few \textit{support inputs} and corresponding
\textit{support outputs} for a given \textit{query} sequence are provided and the task is to predict
the correct \textit{target} sequence \cite{conf/nips/Lake19, conf/cogsci/LakeLB19, conf/acl/ConklinWST20}.
This has been popularized by the notion of ICL
in large language models, where a few examples of the input-output pairs as well as a query are
given as part of a \textit{prompt}, then the target is predicted autoregressively \cite{conf/nips/BrownMRSKDNSSAA20, conf/naacl/MinLZH22}, which has also been shown to enable
compositional generalization in sequence generation \citep{conf/acl/ChenZZK022, journals/corr/abs-2012-09543/Logeswaran/2020}.

\subsection{Retrieval Methods for In-Context Learning}
ICL methods are sensitive to the choice of support sets used.
\citet{conf/aaai/MitchellFM21} found that selecting supports that
were not relevant to the task at hand degraded performance when
using sequence based meta-learning with SCAN.
As we also show in our experiments, ICL
 approachs with a poorly chosen procedure for selecting supports
may be worse on all tasks compared to when no ICL is used at all.

Different approaches have been proposed for finding good examples.
One approach is to try to "tune" the supports directly, either with a
gradient based method \citep{conf/emnlp/LesterAC21, conf/emnlp/ShinRLWS20} or 
by reinforcement learning \citep{conf/emnlp/Deng22}.
Such methods are theoretically attractive, but are difficult optimization
problems to solve in absence of appropriate validation data. Other methods try to pick good examples from the training data,
for example by using a similarity index \citep{conf/emnlp/PasupatZG21},
or with a metric that takes into account diversity and local structure
coverage \cite{journals/corr/abs-2212-06800/Levy/2022, journals/corr/abs-2305-14907}. \citet{conf/coling/Zemlyanskiy22} generates a possible output
candidate for the query input, then searches
the training data for similar outputs, but this depends on a good
initial generation of the output, in the sense that it should be close in the 
output space to useful supports. Retrieval based approaches
all have the same drawback on a task like gSCAN however, which is that the optimal
supports inputs and outputs for some test splits simply may not exist in the training data. In gSCAN,
we found that most states don't have very close near neighbours (Appendix \ref{sec:gscan_nn_similarity_distribution}), so demonstrations from the training data
might not be in exactly the same state.

Closer to this work are generative approaches, for example subproblem decomposition \citep{conf/naacl/Yang22}, chain-of-thought \cite{conf/neurips/Wei22, journals/corr/abs-2205-11916/Kojima/2022}, least-to-most-prompting \cite{journals/corr/abs-2205-10625/Zhou/2022, journals/corr/abs-2209-15003/Drozdov/2022, journals/corr/abs-2210-03493/Zhang/2022} and asking for diverse examples \cite{journals/corr/abs-2305-15035/Chen/2023, conf/iclr/0002IWXJ000023}. These approaches can get very impressive results on ungrounded compositional
generalization benchmarks, but they have their own requirements including reliance on information in large language models, special helper prompts about the input structure. Our work extends on the generated-example paradigm with the idea of generating support instructions for a query state, then solving those support instructions using a model. We explain later in Section \ref{sec:support_gen} why this is particularly important in the grounded language learning setting.

\subsection{Compositional Generalization and Grounded Language Learning}

The capability of Deep Learning to perform compositional generalization has been studied
extensively. Early experiments showed the challenge of doing so on both RNNs \cite{conf/icml/LakeB18} 
and Transformers \cite{journals/jair/HupkesDMB20} and many datasets have been created to
demonstrate the problem, both with synthetic and ``realistic" natural language data \cite{conf/emnlp/BastingsBWCK18, conf/emnlp/KimL20, conf/iclr/KeysersSSBFKMSS20, conf/acl/LiYCZ20, conf/naacl/YinFNPPSTA21, conf/acl/RadevKZZFRS18}. As more datasets become
available, so do approaches to handle the compositional generalization problem. Most approaches
generally fall into some combination of data augmentation \citep{conf/acl/Andreas20, conf/neurips/Li22, journals/corr/abs-2208-10722/Chen/2022, conf/naacl/QiuSPNLST22, conf/iclr/Akyurek21}, neural module networks \citep{conf/cvpr/2016/AndreasRDK15, conf/TACL/Buch2021, conf/nips/DAmarioSB21, conf/naacl/AndreasRDK16, conf/cogsci/Ruis22} and meta-learning \citep{conf/nips/Lake19, conf/acl/ConklinWST20}, discussed in more detail in the next section.

Compositional generalization is also a highly relevant problem in the field of autonomous agents
and robotics as well. In robotics there is typically a richer observation space and it has
been shown that some level of compositional generalization is possible when it comes to manipulating
unseen objects or objects in novel ways \cite{conf/corl/JangIKKELLF21, journals/corr/abs-2106-02972/Goyal/2021, conf/iclr/HillLSCBMS20, conf/neurips/Garg22}, but the success rates are still below a level that could be considered reliable.

Language grounded agents (often referred to as ``Grounded Language Learning" agents) are a natural
fit to study this problem, because it is easy to test compositional generalization scenarios by
varying the input utterance composition and checking if a corresponding composition of actions is executed
by the agent. The most relevant environment for studying
compositional generalization in Grounded Language Learning is gSCAN \cite{conf/nips/RuisABBL20}\footnote{MIT License \url{github.com/LauraRuis/groundedSCAN}},
which has a single training data set and 8 out-of-distribution test splits
covering various compositional generalization scenarios.

gSCAN is a Minigrid-based environment where an agent receives an instruction with a target
object, a verb to apply to that object and an adverb which affects both navigation and the verb.
About 360,000 demonstrations of navigating to various objects and performing some task on them with
various adverbs are provided as a training set. A \textit{success} happens when the agent performs
the expected sequence of actions exactly. The input and action vocabularies are small and the
instructions constructed using a simple grammar. Typically the instructions follow the form
``[verb] a [size] [color] [object] [adverb]", where [size], [color] and [adverb] are sometimes
omitted. The in-distribution split is 100\% solvable by deep learning. More challenging are the eight out-of-distribution test splits. The splits
can be categorized into two categories. The first category, splits B, C, E, F require a
compositional understanding of the input to identify the goal object, for example identifying a ``red square" as a goal
in split C and a size-3 object being ``small" in relation to other objects in split E. Extensions to gSCAN such as ReaSCAN \cite{conf/neurips/Wu21} and Relational Splits (gSCAN-RS) \cite{conf/emnlp/QiuH0SS21} test further such scenarios. The second category, splits D, G, H and I, require
entirely new outputs to be produced at testing-time. Split D requires navigating
to an object that is south-west of the agent, which in practice requires the production of
\texttt{LTURN(3)}\footnote{In this work, where an action or subsequence is repeated $n$ times, we use the notation \texttt{(ACT1 ACT2)(n)}}. Split H requires composing a the verb ``pull" with the adverb ``while spinning",
which requires the production of novel fragments \texttt{LTURN(4)} \texttt{PULL}. Split G is a few-shot learning split.

Various approaches to gSCAN including graph networks \citep{conf/ijcnlp/GaoHM20}, linguistic-assisted attention \citep{conf/emnlp/KuoKB21}, symbolic reasoning \cite{conf/nips/Nye21}, auxiliary tasks \cite{conf/emnlp/JiangB21, conf/blackboxnlp/HeinD22}, modular
networks \citep{journals/corr/abs-2009-13962/Heinze-Deml/2020, conf/cogsci/Ruis22}, logic programming \cite{conf/acl2023/yang23} and data augmentation \citep{journals/corr/abs-2201-11766/Setzler/2022, conf/cogsci/Ruis22} have been proposed. These
approaches tend to make some trade-off between performance and generalizability. 
Transformers have been shown to work well on on the first category of splits \citep{conf/emnlp/QiuH0SS21} as well as on ReaSCAN and gSCAN-RS \citep{conf/emnlp/Sikarwar22}, but
there is no general approach which works well on the second category. In this work, we aim to show
that an ICL approach along with a support generation strategy that does not assume too
much about the problem is a feasible general approach at least for problems like the one in Split H.

\section{Method}

In this section, we describe our implementation of DemoGen. The method is designed to work with datasets like gSCAN where there is both an instruction and a state in the input.

\subsection{In-context-learning}
\label{sec:meta-learning}

Our ICL architecture is a large-context encoder-decoder Transformer (see Fig.~\ref{fig:transformer}).
For a given episode with the initial state $S$ and instruction $I_q$, the model is trained to generate a sequence of targets $A^Q=a^Q_1, ..., a^Q_{m}$ using a set of \textit{support inputs} $I_1, ..., I_n$ and the corresponding \textit{support outputs} $A_1, ..., A_n$.

The entire set of support states $S_1, ..., S_n$, support
instructions $I_1, ..., I_n$ and corresponding support targets $A_1, ..., A_n$, along with the query state
$S_q$ and query instruction $I_q$ are passed as one big
sequence to the Transformer Encoder, using sine-cosine
positional encoding in \cite{conf/nips/VaswaniSPUJGKP17}.
Right-shifted query targets are used as inputs to the
Transformer Decoder with causal masking.
{\color{black}One key difference, similar to \textbf{metaseq2seq} \cite{conf/nips/Lake19} is that both the support targets
and query targets passed through a \textit{permutation} step,
where the symbol-index mapping is different for each data
point. This helps to prevent overfitting and forces the
model to use the supports to produce the correct outputs.} For example, the sequence "WALK(5) RTURN WALK(5)" would be translated into "RTURN(5) LTURN RTURN(5)" under the permutation WALK $\rightarrow$ RTURN, RTURN $\rightarrow$ LTURN. \textcolor{black}{It is possible that a \textit{query target} with the same symbols for \texttt{pull} ... \texttt{while spinning}
is generated after permutation during training, however the probability of this happening is low. We measured that for a single pass
through the training data, approximately 3\% of the generated query instructions matched \texttt{pull} ... \texttt{while spinning}, 0.3\%
of the permuted query outputs matched \texttt{PULL} actions followed by four \texttt{LTURN} instructions, and their intersection was
0.001\% of all sampled supports.} A complete table showing the effect of different permutations is shown in Appendix \ref{sec:appendix_permuter}.

\subsection{Support Set Generation}
\label{sec:support_gen}
Choosing the support inputs $I_1, ..., I_n$ and outputs $A_1, ..., A_n$ for the ICL model is not a trivial problem. DemoGen generates the support sets using generative models trained on the training data, similar to the idea in \citet{journals/corr/abs-2305-15035/Chen/2023} for ungrounded language compositional generalization.

Support inputs are generated by an encoder-decoder language masked language model, similar to BART \cite{conf/acl/LewisLGGMLSZ20}. The model is trained to estimate $p(w_0, ..., w_n|w_{j \not \in \mathcal{M}})$ -- reconstructing the input sentence given some non-masked tokens $\mathcal{M}$. The model
is trained on a balanced dataset of all the instructions in the training
data to ensure that inputs occurring less often have a reasonable chance of being sampled.
To generate support inputs, some percentage of the tokens (including padding tokens) in the query $I_q$ (in this work, 20\%) are randomly masked and then instructions are sampled by autoregressive decoding.
This process is repeated $k \ge n$ times, to form $I_1, ..., I_k$.
We deduplicate the samples and remove $I_q$ from $I_1, ..., I_k$.
We also \textit{filter} the supports by the use of a \textit{scoring model}. The scoring model estimates probability that a generated support is in-distribution, conditioned on any relevant context. We assume that conditionally in-distribution supports are more likely to be solveable by the model. {\color{black} A simple solution is to estimate the length-normalized log-likelihood of each of the generated support instructions through the same generation model, which is what we do in this case.}. Then we take the top $n$ by score to get $I_1, ..., I_n$. Finally,
{\color{black} support outputs $A_1, ..., A_n$ are generated from the state-$I_1, ..., I_k$ pairs by a Transformer model pre-trained on the gSCAN training set}. Examples of the generated instructions are shown in Fig.~\ref{fig:oracle} and also in Appendix \ref{sec:generated_demonstrations}.

Generating both the support inputs and outputs has a few interesting advantages. {\color{black}Compared to retrieving on inputs, we can generate examples which we know will be relevant to the current state and also generate examples which might not be found in the training data for a given query (for example, ``red square"). Compared to retrieving based on the predicted output, we can generate a greater diversity of supports which would be valid in the state, as opposed to fetching the same output over and over again in many different states. The only assumption we make is that the
model used to generate the supports generalizes sufficiently to reach different kinds of targets,
but not compositionally to different behaviours for reaching them. In practice,
this is already true with the Transformer architecture \citep{conf/emnlp/QiuH0SS21, conf/emnlp/Sikarwar22}. One challenge with generating the supports is that our support generator might come up with support inputs that are either not relevant or not solvable in the current state.
We show in the experiments that the presence of irrelevant supports is not a problem as long as the other useful supports are also present.

\section{Experiments}

To validate the effectiveness of DemoGen for the grounded
compositional generalization challenge, we evaluate on gSCAN
with a range of different ICL configurations. In all
experiments, we generate up to {\color{black}16} supports for each example in the
training and test splits using the methods described below, then train
the ICL Transformer on the training set augmented with the generated supports,
then evaluate on the test splits augmented with the generated supports.
Examples of each are given in Appendix \ref{sec:generated_demonstrations}. The ICL Transformer is a standard Transformer with 12 encoder and decoder layers and 8 attention heads, with an embedding dimension of 512. Additional hyperparameters
are described in Table \ref{tab:hyperparameters} in the Appendix.

\begin{table}[t]
\centering
\begin{tabular}{l|rrrrrr}
\hline
{}  &  \footnotesize{\textbf{DG}} &  \footnotesize{\textbf{GR}} & \footnotesize{\textbf{CR}}  & \footnotesize{\textbf{OS}} & \footnotesize{\textbf{RD}}  \\
\hline
\footnotesize{(1) Desc. Obj.} & \footnotesize{0.33} & \footnotesize{0.68} & \footnotesize{0.33} & \footnotesize{1.00} & \footnotesize{0.16} \\
\footnotesize{(2) Agent Pos.} & \footnotesize{1.00} & \footnotesize{0.08} & \footnotesize{1.00} & \footnotesize{0.03} & \footnotesize{1.00} \\
\footnotesize{(3) Tgt. Pos.} & \footnotesize{0.44} & \footnotesize{0.08} & \footnotesize{0.39} & \footnotesize{0.03} & \footnotesize{0.16} \\
\footnotesize{(4) Same Diff.} & \footnotesize{0.44} & \footnotesize{0.09} & \footnotesize{0.39} & \footnotesize{0.02} & \footnotesize{0.16} \\
\footnotesize{(5) Tgt. Obj.} & \footnotesize{0.44} & \footnotesize{0.14} & \footnotesize{0.27} & \footnotesize{0.19} & \footnotesize{0.16} \\
\footnotesize{(6) Verb \& (5)} & \footnotesize{1.00} & \footnotesize{0.15} & \footnotesize{0.88} & \footnotesize{0.43} & \footnotesize{0.16} \\
\footnotesize{(7) Advb \& (5)} & \footnotesize{0.88} & \footnotesize{0.51} & \footnotesize{0.78} & \footnotesize{0.33} & \footnotesize{0.16} \\
\footnotesize{(8) (6) \& (7)} & \footnotesize{0.88} & \footnotesize{0.00} & \footnotesize{0.70} & \footnotesize{0.19} & \footnotesize{0.16} \\
\footnotesize{(9) (4) \& (8)} & \footnotesize{0.88} & \footnotesize{0.00} & \footnotesize{0.62} & \footnotesize{0.00} & \footnotesize{0.16} \\
\hline
\end{tabular}
\caption{Analysis of generations over synthetic data, Split H. {\color{black}Not shown is \textbf{Heuristic}, which gets 1.00 in every category by its defintion.}}
\label{tab:analysis_b}
\end{table}

\paragraph{DemoGen (DG)} Our generation strategy as described in Section~\ref{sec:support_gen}. 2048 instructions are sampled from the language model, deduplicated, and ranked to get the top 16 instructions and corresponding support targets for the query state.

\paragraph{Coverage Retrieval (CR, CovR)} Supports are retrieved using a similarity index on states and chosen greedily such that all the one-grams and two-grams in the query instruction are covered, similar to the Set-BSR method in \cite{journals/corr/abs-2305-14907}. The method tries to find similar states with hopefully relevant instructions. See Appendix \ref{appendix:covr} for more details on how Set-BSR was adapted for gSCAN.

\paragraph{GandR (GR)} Supports are retrieved using the Generate-and-Retrieve
strategy \citep{conf/coling/Zemlyanskiy22}.
In this method a vector similarity index of input and target pairs is
built, where the input-output pairs are encoded using TF-IDF. Query ``outputs"
for the test data points come from initial preditions made by a Transformer model. We also extend GandR to greedily pick examples covering the query input to avoid picking the same instruction many times. See Appendix \ref{appendix:gandr} for more details.

\paragraph{Heuristic} An expert generates all valid input and output pairs for a given state and selects the best ones by; 1)~going to the same object, 2)~showing the target verb in combination with other adverbs, 3)~showing the target adverb in combination with other verbs.
Note that the generated supports might contain test-set input-output pairs, meaning that we assume extra knowledge not available to the learning agent. The heuristic can be seen as an upper bound on performance we could expect from an optimal demonstration generator.
See Appendix \ref{sec:appendix_oracle} for more details.

\begin{table}[t]
\centering
\begin{tabular}{l|cccc}
\hline
{} & \textbf{V} &  \textbf{C}  & \textbf{C \& V} & \textbf{C | V}\\
\hline
A &   0.79 &     0.70 &               0.70 &              0.88 \\
B &   0.73 &     0.64 &               0.64 &              0.88 \\
C &   0.61 &     0.50 &               0.50 &              0.83 \\
D &   0.65 &     0.24 &               0.24 &              0.36 \\
E &   0.78 &     0.66 &               0.66 &              0.84 \\
F &   0.73 &     0.63 &               0.63 &              0.87 \\
G &   0.79 &     0.72 &               0.72 &              0.91 \\
H &   0.79 &     0.56 &               0.56 &              0.71 \\
\hline
\end{tabular}
\caption{DemoGen supports, (V)alid instructions, (C)orrect targets, correct and valid (C \& V) and correct given valid (C | V) on synthetic data by split, according to an oracle function.}
\label{tab:correctly_generated}
\end{table}

\paragraph{Rand. Instrs (RD) } The same expert is used but the support instructions are selected randomly, without the use of the heuristic described above. Thus, instructions can
be about any object in the same state, not just the target one.

\paragraph{Other States (OS)} We generate instructions as in the Heuristic approach but demonstrations are in states different to the query state. Such states are extracted from the training data. The sampled states are also included in the supports and used by the ICL Transformer. If the training data does not contain a state with the same instruction as the one generated by the expert, that instruction is not included in the support set.

We also compare against two simpler non-ICL baselines:

\paragraph{Transformer} An encoder-decoder Transformer of the
same configuration as the ICL Transformer, but without any
in-context learning, similar to \citep{conf/emnlp/QiuH0SS21},
but without the initial convolutional and early cross-attention layers.

\paragraph{Fine-Tuning} The same \textbf{Transformer} but including the generated supports for \textbf{DemoGen} in the training data set.

\subsection{Analysis of Generated Instructions}

We analyze some properties of the generated support sets
under different generation conditions for Split H in Table \ref{tab:analysis_b}
(similar analysis for other splits can be found in Appendix \ref{sec:generated_demonstrations_other_splits}).
In retrieval-based methods, the distance between the agent
and the target object is often different in the query versus the supports (4).
Retrieval based methods tend to generate fewer demonstrations showing
the same exact same target object (5). The target object might vary because the instruction
can be under-specified (for example, ``walk to a square", where the only square
in the query state is a red square, but it would be perfectly valid to fetch an example
where the target was a blue square). Retrieval methods
do not always have both (8) the correct verb (6) and adverb (7) in the
retrieved supports. This happens on GandR because the adverb can quite significantly
change the outputs, such that supports with the same verb (but without the 
adverb) are not selected. In even fewer cases will there be at least one demonstration
each  of both the correct verb and adverb on a trajectory covering the same path as the 
one in the query (9).  Our method on the other hand is able to to generate demonstrations
which do have these properties.

One important question for our method is the quality of
the generated supports. Ideally they should comprise \textit{valid}
support inputs (eg, tasks that are actually solveable in a state)
and the generated support outputs should be \textit{correct}
enough to facilitate ICL. We investigated this
on supports generated by our method and reported the results
in Table \ref{tab:correctly_generated}. On average, about
77\% of generated support inputs are valid. A support output
is \textit{correct} if it matches what an oracle generator
would have generated for the corresponding instruction and
state. 50\% of the support pairs
were both correct and valid. The
number is clearly lower on splits where a Transformer is not
able to solve the task well. For example on Split H, there may be ``pull an [object] while spinning" in the generated support inputs, where [object] is not the target object.

%\section{Discussion and Analysis}

\subsection{Performance on gSCAN}
\label{sec:perf}

\begin{table}[t]
\centering
%\resizebox{\hsize}{!}{
\begin{tabular}{lllll}
\hline
 & TF & CovR & GandR & DemoG \\
\hline
A & \textbf{1.0} & 0.99 \footnotesize{± 0.01} & 0.99 \footnotesize{± 0.01} & 1.0 \footnotesize{± 0.01} \\
B & \textbf{1.0} & 0.98 \footnotesize{± 0.01} & 0.88 \footnotesize{± 0.05} & 1.0 \footnotesize{± 0.01} \\
C & 0.96 & 0.83 ± 0.3 & 0.92 \footnotesize{± 0.03} & 0.98 \footnotesize{± 0.02} \\
D & 0.01 & 0.0 ± 0.0 & 0.0 ± 0.0 & 0.03 \footnotesize{± 0.02} \\
E & 0.9  & 0.99 \footnotesize{± 0.01} & 0.99 \footnotesize{± 0.01} & 0.99 \footnotesize{± 0.01} \\
F & \textbf{1.0}  & 0.99 \footnotesize{± 0.01} & 0.99 \footnotesize{± 0.01} & 0.99 \footnotesize{± 0.01} \\
G & 0.0 & 0.0 ± 0.0 & 0.0 ± 0.0 & 0.0 ± 0.0 \\
H & 0.22 & 0.56 ± 0.1 & 0.17 \footnotesize{± 0.01} & 0.8 \footnotesize{± 0.05} \\
\hline
\end{tabular}
\iffalse
\begin{tabular}{c|l|ll}
\hline
{} &  \multicolumn{1}{c}{Transformer} &      \multicolumn{1}{c}{GandR} &        \multicolumn{1}{c}{DemoGen} \\
\hline
A & \textbf{1.0} \footnotesize{± 0.0} & 0.99 \footnotesize{± 0.0} & 1.0 \footnotesize{± 0.01} \\
B & \textbf{1.0} \footnotesize{± 0.0} & 0.9 \footnotesize{± 0.09} & 1.0 \footnotesize{± 0.01} \\
C & 0.96 \footnotesize{± 0.06} & 0.94 \footnotesize{± 0.04} & 0.98 \footnotesize{± 0.02} \\
D & 0.01 \footnotesize{± 0.02} & 0.0 \footnotesize{± 0.0} & 0.03 \footnotesize{± 0.02} \\
E & 0.9 \footnotesize{± 0.19} & 0.97 \footnotesize{± 0.0} & 0.99 \footnotesize{± 0.01} \\
F & \textbf{1.0} \footnotesize{± 0.0} & 0.96 \footnotesize{± 0.01} & 0.99 \footnotesize{± 0.01} \\
G & 0.0 \footnotesize{± 0.0} & 0.0 \footnotesize{± 0.0} & 0.0 \footnotesize{± 0.0} \\
H & 0.22 \footnotesize{± 0.0} & 0.08 \footnotesize{± 0.02} & \textbf{0.8} \footnotesize{± 0.05} \\
\hline
\end{tabular}
\fi
%}
\caption{Success rates for different splits (A--H). Numbers are ± standard deviation over 10 seeds, measured after 300,000 steps. {\color{black}CovR, GandR and DemoGen (ours) both use the ICL Transformer as the model architecture, with supports generated by each method. Best results bolded. }}
\label{tab:results}
\end{table}

In Table~\ref{tab:results} we show the results of evaluation on gSCAN for DemoGen, closely related baselines and simple baselines. For all experiments, training was run for 300,000 iterations then take the best model checkpoints on the in-distribution Split-A validation loss,
and evaluate on all the other splits.

The Transformer can perform very well on the in-distribution
Split A, and as expected, performance on splits B, C, E and F is
also very good. But performance on Split H is poor.
Simple architectural changes like Rotary Positional Encodings or the Universal Transformer also do not make much of a difference.
More comparisons to other non-ICL prior work on gSCAN are in Appendix \ref{sec:additional_comparisons}.

ICL methods can perform much better on Split H. Coverage Retrieval is a strong baseline that gets a success rate of 56\%, but has high variance between seeds Split C in particular. GandR gets 17\% on this split, but retains good performance on the other split. Upon inspection of the type of demonstrations that GandR fetches, we notice that it mostly retrieved demonstrations concerning the adverb, or the verb, but not
both types into the same demonstration set. This can be seen in the fetched trajectories
example in Appendix \ref{sec:appendix_gen_data}. Our method, DemoGen, gets 80\% on average,
and retains good performance on the other splits as well. Notably,
generation of supports fares better than mere retrieval, even if we
try to retrieve examples that are very close to the query state, which is what CovR tries to do.

On Splits D and G, performance is still 
not good. The reason is they require generation of a pattern that won’t be seen in the outputs in any permutation of the labels. In the case of
Split D, it requires LTURN(2) WALK(n) LTURN(1) WALK(n). Only 6\%
of the data matches this pattern in any index-label permutation. In
the case of split G, (LTURN RTURN(3) LTURN WALK)(n) is required. Only
0.0001\% matches that up to a permutation. In contrast, Split H requires
(LTURN(4) PULL(n)), and there are many examples from the ``push a
[size] [color] [object]`` set of instructions matching that up to a
permutation.

\begin{table}[t]
\centering
\begin{tabular}{llll}
\hline
{} &    \multicolumn{1}{c}{Heuristic} &       \multicolumn{1}{c}{Rand. Instrs} & \multicolumn{1}{c}{Other States} \\
\hline
A     &    1.0 \footnotesize{± 0.0} &  0.77 \footnotesize{± 0.01} &   0.99 \footnotesize{± 0.0} \\
B     &    1.0 \footnotesize{± 0.0} &  0.62 \footnotesize{± 0.21} &    0.0 \footnotesize{± 0.0} \\
C     &    1.0 \footnotesize{± 0.0} &  0.66 \footnotesize{± 0.11} &    0.2 \footnotesize{± 0.0} \\
D     &   0.5 \footnotesize{± 0.07} &    0.0 \footnotesize{± 0.0} &    0.0 \footnotesize{± 0.0} \\
E     &    1.0 \footnotesize{± 0.0} &   0.59 \footnotesize{± 0.1} &    0.0 \footnotesize{± 0.0} \\
F     &    1.0 \footnotesize{± 0.0} &  0.75 \footnotesize{± 0.05} &  0.99 \footnotesize{± 0.01} \\
G     &    0.0 \footnotesize{± 0.0} &    0.0 \footnotesize{± 0.0} &    0.0 \footnotesize{± 0.0} \\
H     &  0.86 \footnotesize{± 0.03} &  0.15 \footnotesize{± 0.02} &   0.0 \footnotesize{± 0.01} \\
\hline
\end{tabular}
\caption{
Success rates for different types of different types of oracle behaviour. Numbers are success rates ± standard deviation with
the same measurement methodology as Table \ref{tab:results}}
\label{tab:oracle_study}
\end{table}

\subsection{Architectural Ablations}
\label{sec:arch_ablations}

\begin{table}
\centering
\begin{tabular}{l|cc}
\hline
 & Fine-Tuning & No Permutations \\
\hline
A & 1.0\footnotesize{ ± 0.0} & 0.94\footnotesize{ ± 0.0}6 \\
B & 1.0\footnotesize{ ± 0.0} & 0.92\footnotesize{ ± 0.0}5 \\
C & 1.0\footnotesize{ ± 0.0} & 0.72\footnotesize{ ± 0.2}8 \\
D & 0.16\footnotesize{ ± 0.0}1 & 0.0\footnotesize{ ± 0.0} \\
E & 1.0\footnotesize{ ± 0.0} & 0.92\footnotesize{ ± 0.0}9 \\
F & 1.0\footnotesize{ ± 0.0} & 0.92\footnotesize{ ± 0.0}8 \\
G & 0.0\footnotesize{ ± 0.0} & 0.0\footnotesize{ ± 0.0} \\
H & 0.22\footnotesize{ ± 0.0} & 0.18\footnotesize{ ± 0.0}2 \\
\hline
\end{tabular}
\caption{Other architectural changes, evaluated on 5 random seeds on best split-A performance.}
\label{tab:arch_changes}
\end{table}

We also compare other architectural changes
to validate our results in Table \ref{tab:arch_changes}. Fine-tuning on the generated data can improve Split D performance, but not Split H performance. Removing the permuter block (No Permutatons) makes DemoGen's Split H performance worse, on a similar level to not using ICL at all, similar to
what was reported in \citet{conf/nips/Lake19}. If we remove the filtering of demonstrations and instead only take the first 16 generated demonstrations, then ....

\subsection{Comparing Retrieval Oracles}
\label{sec:analysis}

In Table \ref{tab:oracle_study}, we analyze the importance of the strategy used to
select the support sets by evaluating the performance of
three hand-written oracle functions on the ICL Transformer. \textbf{Heuristic} gets
very high scores, since it samples only the instructions
and actions known a-priori to be relevant the query instruction.
However, without care in sampling the supports, performance
drops significantly on all-splits, including the in-distribution
ones. For example, \textbf{Random} instruction sampling yields instructions
irrelevant to the query task (because they concern different objects), leading to bad performance on all splits. Retrieving demos in \textbf{Other States} for known good instructions is even worse.
In some splits, it is not possible to sample from the training
data as there is no example of an instruction concerning the
same object as in the query.

\subsection{Importance of Good Demonstrations}
\label{sec:good-demonstrations}

\begin{figure}[ht]
\centering
\includegraphics[width=\columnwidth, keepaspectratio=1]{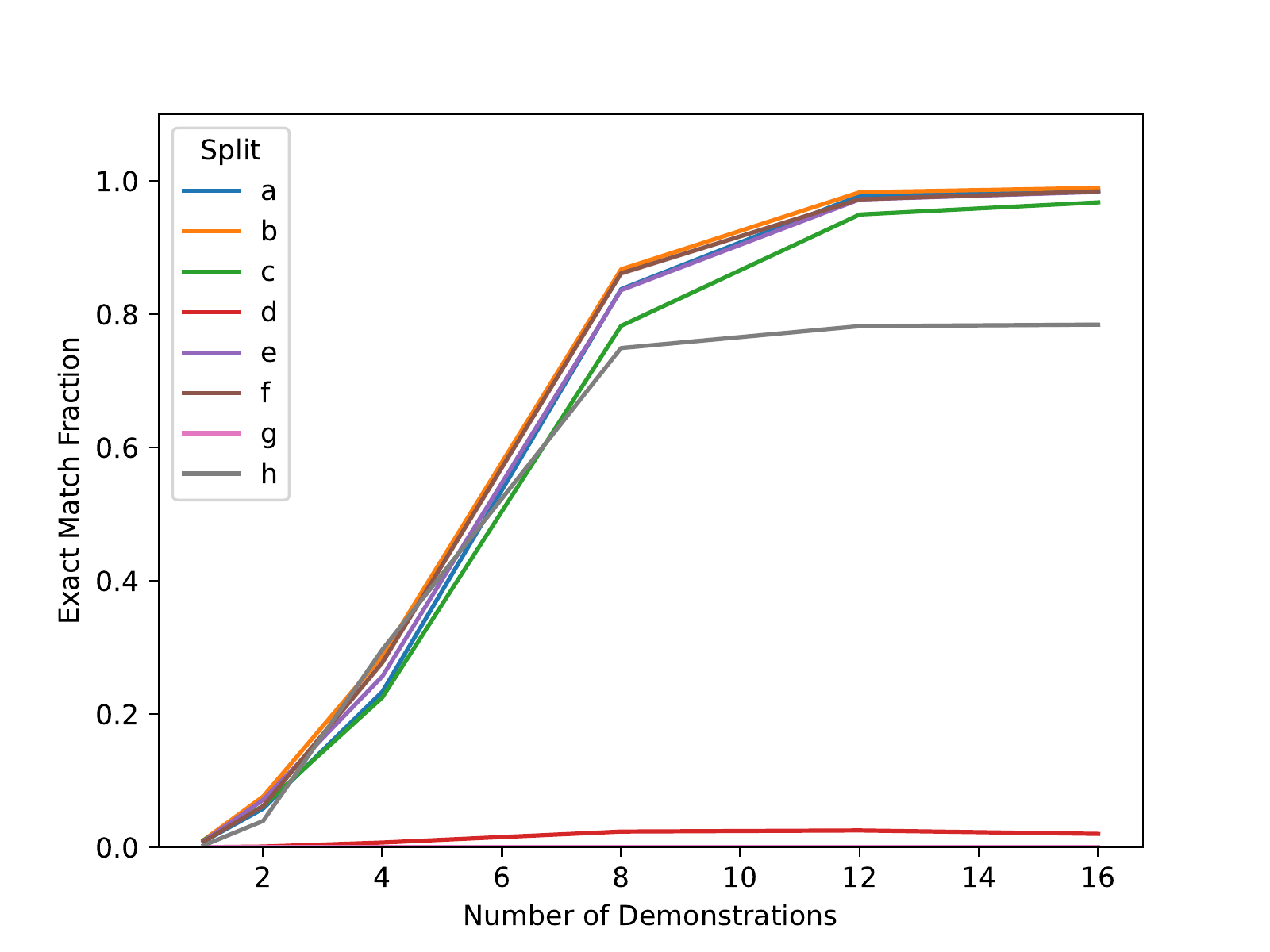}
\caption{Performance of a DemoGen trained ICL Transformer with different numbers of demonstrations at evaluation time on each split. Performance was evaluated over models trained with 10 different seeds.}
\end{figure}

\subsection{Scaling to natural language}
\label{sec:scaling}

{\color{black}One criticism of the gSCAN dataset is that because it is synthetically
generated, good results on gSCAN may not generalize to instructions that resemble
something more like natural language. To test the robustness of our method in this setting, we generate a new derivative of gSCAN called NL-gSCAN by
using a large language model to generate paraphrases of the instructions. By prompting the \texttt{openai-gpt3.5} model with 25 different examples of paraphrases for an instruction, we can
generate paraphrases of all the other instructions in the dataset. The word frequency
distribution more closely matches a Zipf distribution, and there is a greater diversity
of both words and syntax parses. Information about the target object was retained in approximately 99\% of cases. Further details are given in Appendices \ref{sec:appendix_gen_data} and \ref{sec:appendix_gen_data_props}. The sentence structure, adverbs and verbs may be written in different ways for different objects. We also evaluated on an image-based dataset, the details of that dataset and our evaluation can be found in Appendix \ref{sec:appendix_gscan_img}.

\begin{table}[t]
\centering
\setlength{\tabcolsep}{3pt}
% Note: Not yet complete
\begin{tabular}{c|llll}
\hline
{} & \multicolumn{1}{c}{TF} & \multicolumn{1}{c}{DG} & \multicolumn{1}{c}{CR} & \multicolumn{1}{c}{GR} \\
\hline
A & \footnotesize{1.0} \footnotesize{± 0.0} & \footnotesize{0.99} \footnotesize{± 0.0} & \footnotesize{0.96 ± 0.07} & \footnotesize{0.94 ± 0.02} \\
B & \footnotesize{0.99} \footnotesize{± 0.0}  & \footnotesize{0.96} \footnotesize{± 0.0} & \footnotesize{0.91 ± 0.08} & \footnotesize{0.9 ± 0.06} \\
C & \footnotesize{0.99} \footnotesize{± 0.03} & \footnotesize{0.97} \footnotesize{± 0.0} & \footnotesize{0.54 ± 0.3} & \footnotesize{0.84 ± 0.07} \\
D & \footnotesize{0.08} \footnotesize{± 0.16}  & \footnotesize{0.01} \footnotesize{± 0.01} & \footnotesize{0.0 ± 0.0} & \footnotesize{0.0 ± 0.0} \\
E & \footnotesize{0.98} \footnotesize{± 0.03} & \footnotesize{0.98} \footnotesize{± 0.0} & \footnotesize{0.96 ± 0.06} & \footnotesize{0.9 ± 0.04} \\
F & \footnotesize{1.0} \footnotesize{± 0.0} & \footnotesize{0.98} \footnotesize{± 0.0} & \footnotesize{0.86 ± 0.11} & \footnotesize{0.94 ± 0.02} \\
G & \footnotesize{0.0} \footnotesize{± 0.0} & \footnotesize{0.0} \footnotesize{± 0.0} & \footnotesize{0.0 ± 0.0} & \footnotesize{0.0 ± 0.0}  \\
H & \footnotesize{0.19} \footnotesize{± 0.03} & \footnotesize{0.59} \footnotesize{± 0.06} & \footnotesize{0.43 ± 0.12} & \footnotesize{0.18 ± 0.03} \\
\hline
\end{tabular}
\caption{Evaluation of baselines and DemoGen on a paraphrased version of gSCAN. Numbers are fraction of exact matches.}
\label{}
\end{table}

After generating a paraphrased dataset, a DemoGen dataset can be generated in exactly the same way
as specified in Section \ref{sec:support_gen}. The Transformer baseline is able to solve the other splits reasonably well on this new dataset. DemoGen maintains good
performance compared to a Transformer and still gives a 40 point a performance boost on Split H, though
the improvement is not as substantial compared to the synthetic data.}

\section{Conclusion}
\label{sec:conclusion}

{\color{black}In this work we examined a case where it was necessary to generate
support sets for ICL in a grounded language learning problem.
We proposed a method for doing so based on sampling from
an autoregressive language model and solving the generated support inputs
using a transformer trained on the training data. Our method performs
well on many of the gSCAN splits, including the challenging Split H. {\color{black}The
method also scales well to instructions resembling natural language.}
We analyze the nature of the generated supports and show that
they contain useful information are typically valid and correct.}

\clearpage

\section{Limitations}
\label{sec:limitation}

{\color{black}In this section, we discuss the limitations of our work.

First, on the dataset and evaluation. gSCAN is a synthetic and with quite
simple instructions. We wanted to evaluate on instructions that were challenging
like natural language, but we did not have the resources to crowdsource annotations
for every data point in gSCAN. Therefore, we relied on commercial large language models to
generate similar instructions instead. These instructions aren't a substitute for exactly
human-generated language, but they are a good approximation.

Another limitation of this work is that supports
need to be generated at test time for the test set.
In this work, we pre-generated the supports for the test
set, though a real-time application of this work on unseen
examples would need to run the generation process,
which could make inference time much longer. There are also
other methods to improve the performance of the support input and support output
procedure, for example quantization \cite{journals/corr/abs-2208-07339/Dettmers/2022}, KV-caching, early stopping, etc.
}

\section{Ethics}
\label{sec:ethics}

{\color{black}We used commercial large language models to generate paraphrases of the
inputs to test the scalability of our method to natural language data in
Section \ref{sec:scaling}. These commercial large language models come with their own
range of documented ethical issues, such as the capability to amplify harmful
biases and misinformation, labour exploitation in training,
energy consumption and permission to use web-scale training
data. There is also an economic ethical aspect, where the use of the large language
model displaces humans who may have been willing to perform the labelling. For our
usecase, it was by many orders of magnitude cheaper to use the large language model
than crowd-sourced labelling at a fair wage. On the other hand, we believe that there
are better uses of human time than paraphrasing hundreds of thousands of examples of
simple navigation problems for the purpose of producing a single research paper.

Our work covers the foundational issue of compositional generalization in
grounded language learning, so we don't expect direct applications of it to
have the potential to cause social harm. However, the work should be adapted
with care. In particular, it is important that the model generating the supports
for ICL is actually generating supports which are useful for generating
the downstream problem. Generating outputs to a problem with generated wrong
input-output pairs is likely to result in even more wrong outputs. Our work shouldn't
be deployed in safety critical situations, but instead should be seen as a step
towards achieving better data-driven compositional generalization.}

\iffalse
\section*{Acknowledgements}

We would like to acknowledge the anonymous reviewers of this paper in its various
submissions, as well as our colleagues Nicola Dainese and Ananth Mahadevan
for their valuable feedback on prior versions of this work.
Computational resources were generously provided by the Aalto Science-IT project
and CSC – IT Center for Science, Finland.
We also acknowledge the the support within the Academy of Finland Flagship programme: Finnish Center for
Artificial Intelligence (FCAI).
\fi

\nocite{*}
\section{Bibliographical References}\label{sec:reference}

\bibliographystyle{lrec-coling2024-natbib}
\bibliography{custom.bib}

\section{Language Resource References}
\label{lr:ref}
\bibliographystylelanguageresource{lrec-coling2024-natbib}
\bibliographylanguageresource{languageresource}

\vfill\null
\pagebreak
\vfill\null
\pagebreak

\appendix

\section{Computational Resource Usage and Reproducibility Requirements}
\label{sec:appendix_resource_usage}

Experiments were run on our internal GPU cluster. Running a ICL
experiment to 300,000 iterations takes about 3 days on a MI250x GPU.
For 6 different experiment runs with 10 seeds each, the total compute time
is about 330 GPU-days, though the experiments can be run in parallel. The number of GPU-days we
used to produce this work was much higher, because of
tweaks to the experimental conditions, debugging, restarting
failed jobs, etc.

\section{Details of the gSCAN Dataset}
\label{appendix:gscan_dataset}

Statistics on the gSCAN dataset are reproduced
in Table \ref{tab:gscan_statistics} for the reader's convenience.

\begin{table}[H]
\centering
\begin{tabular}{lrr}
\hline
{}    & Num. Examples & Length ± std. \\
\hline
Train & 367933 & 14.35 ± 10.07 \\
A & 19282 & 13.35 \footnotesize{± 8.87} \\
B & 18718 & 13.95 \footnotesize{± 9.72} \\
C & 37436 & 14.07 \footnotesize{± 9.78} \\
D & 88642 & 17.96 ± 10.78 \\
E & 16808 & 13.31 \footnotesize{± 9.48} \\
F & 11460 & 16.50 ± 12.40 \\
G & 112880 & 33.46 ± 16.90 \\
H & 38582 & 43.07 ± 19.67 \\
\hline
\end{tabular}
\caption{Statistics on the gSCAN dataset and test splits}
\label{tab:gscan_statistics}
\end{table}

\subsection{Nearest Neighbour Similarity Distribution}
\label{sec:gscan_nn_similarity_distribution}

\begin{figure*}[t]
\includegraphics[keepaspectratio=1, width=\linewidth]{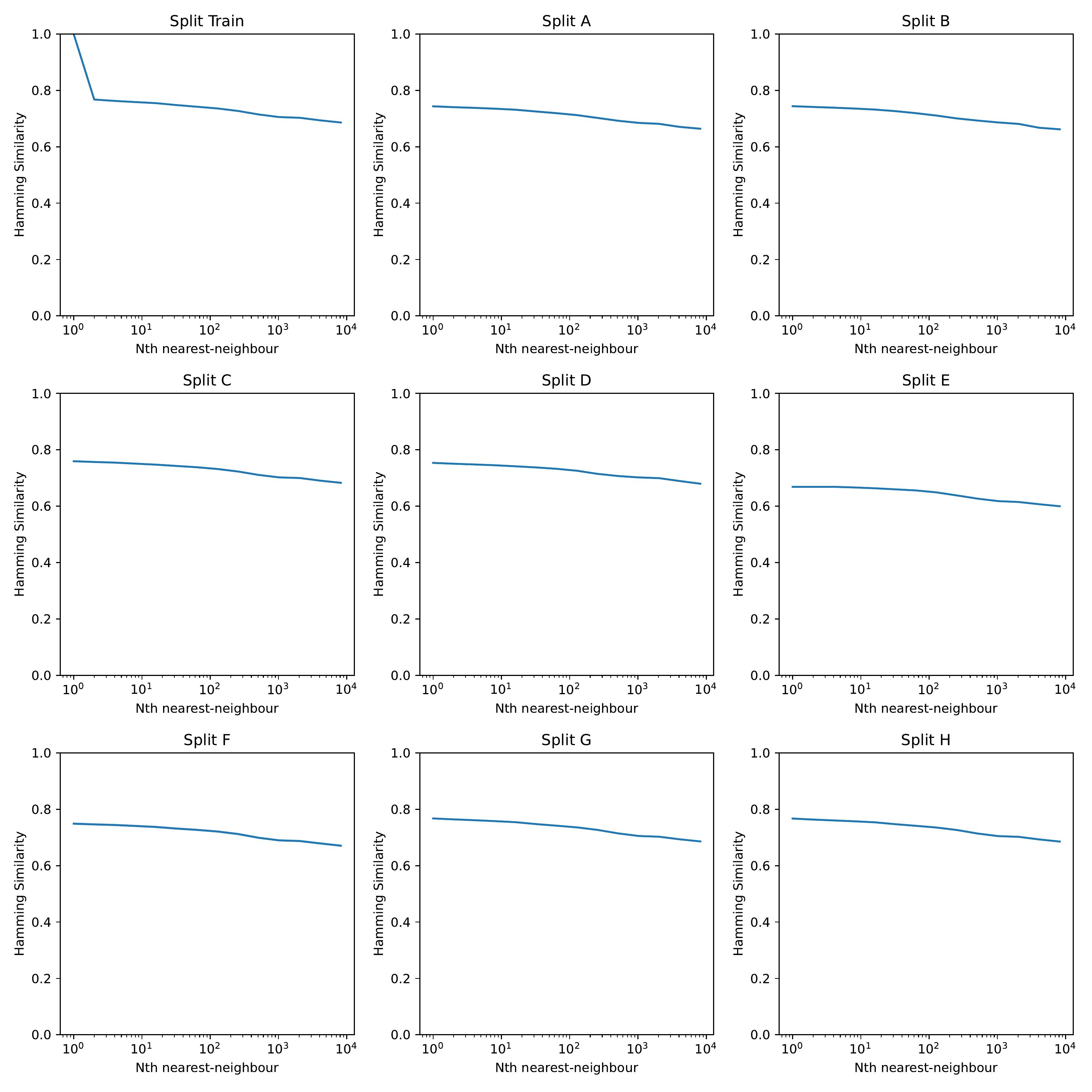}
\caption{Average state nearest neighbour similarity (between the shown split and the training split) for each split. X-axis is log-scale. The plots show the average cosine similarity between points in a split and their Nth nearest
neighbour in the training split.}
\label{fig:avg_nn_similarity}
\end{figure*}

We visualize the average nth training-data nearest neighbour similarity 
distribution for each dataset split in Figure \ref{fig:avg_nn_similarity}.
We created the figure by taking 1000 random examples from each splits, then
finding their 8192 nearest neighbours using a inner-product index over
normalized one-hot encoded state representations.

In most cases, even the closest nearest neighbour state has quite many
differences and these differences grow as we pick nearest neighbours further
away from a training data point. This means that it is hard to find an example in the training set containing different instructions in the exact same environment layout.

\section{Additional Comparisons}
\label{sec:additional_comparisons}

\begin{table*}[ht]
\resizebox{\textwidth}{!}{
\centering
\begin{tabular}{l|cccccc} 
\toprule
              & seq2seq       & GECA  & FiLM       &    RelNet & LCGN & ViLBERT     \\ 
%\hline
              & \citep{conf/nips/RuisABBL20}  & \citep{conf/nips/RuisABBL20}  & \citep{conf/emnlp/QiuH0SS21}  & \citep{conf/emnlp/QiuH0SS21} & \citep{conf/ijcnlp/GaoHM20} & \citep{conf/emnlp/QiuH0SS21}  \\ 
\hline
A             & 97.15 \footnotesize{± 0.46}  & 87.6 \footnotesize{± 1.19}   & 98.83 \footnotesize{± 0.32} & 97.38 \footnotesize{± 0.33} & 98.6 \footnotesize{± 0.9}    & 99.95 \footnotesize{± 0.02}         \\
B             & 30.05 ± 26.76 & 34.92 ± 39.30 & 94.04 \footnotesize{± 7.41} & 49.44 \footnotesize{± 8.19} & 99.08 \footnotesize{± 0.69}  & 99.90 \footnotesize{± 0.06}      \\
C             & 29.79 ± 17.70 & 78.77 \footnotesize{± 6.63}  & 60.12 \footnotesize{± 8.81} & 19.92 \footnotesize{± 9.84} & 80.31 ± 24.51 & 99.25 \footnotesize{± 0.91}    \\
D             & 0.00 \footnotesize{± 0.00}         & 0.00 \footnotesize{± 0.00}   & 0.00 \footnotesize{± 0.00}  & 0.00 \footnotesize{± 0.00}  & 0.16 \footnotesize{± 0.12} & 0.00 \footnotesize{± 0.00}        \\
E             & 37.25 \footnotesize{± 2.85}  & 33.19 \footnotesize{± 3.69}  & 31.64 \footnotesize{± 1.04} & 42.17 \footnotesize{± 6.22} & 87.32 ± 27.38 & 99.02 \footnotesize{± 1.16}     \\
F             & 94.16 \footnotesize{± 1.25}  & 85.99 \footnotesize{± 0.85}  & 86.45 \footnotesize{± 6.67} & 96.59 \footnotesize{± 0.94} & 99.33 \footnotesize{± 0.46}  & 99.98 \footnotesize{± 0.01}   \\
H             & 19.04 \footnotesize{± 4.08}  & 11.83 \footnotesize{± 0.31}  & 11.71 \footnotesize{± 2.34} & 18.26 \footnotesize{± 1.24} & 33.6 ± 20.81  & 22.16 \footnotesize{± 0.01} \\
\hline
              & GroCoT & Planning & RD Random/RL & Modular & CMA-ES & Role-Guided      \\ 
%\hline
              & \cite{conf/emnlp/Sikarwar22} & \citeyear{journals/corr/abs-2009-13962/Heinze-Deml/2020} & \citep{journals/corr/abs-2201-11766/Setzler/2022} & \cite{conf/cogsci/Ruis22} &  \cite{conf/blackboxnlp/HeinD22} & \cite{conf/emnlp/KuoKB21} \\ 
\hline
A             & 99.9    & 94.19 \footnotesize{± 0.71} & 98.39 \footnotesize{± 0.17} & 96.34 \footnotesize{± 0.28} & 99.7 \footnotesize{± 0.1}    & 96.73 \footnotesize{± 0.58}          \\
B             & 99.9    & 87.31 \footnotesize{± 4.38} & 62.19 ± 24.08 & 59.66 ± 23.76 & 73.5 ± 25.4 & 94.91 \footnotesize{± 1.30}      \\
C             & 99.9    & 81.07 ± 10.12 & 56.52 ± 29.70 & 32.09 \footnotesize{± 9.79} & 99.4 \footnotesize{± 0.4}  & 67.72 ± 10.83   \\
D             & 0.0     &               & 43.60 \footnotesize{± 6.05} & 0.00 \footnotesize{± 0.0} & 2.2 \footnotesize{± 1.5}      & 11.52 \footnotesize{± 8.18}   \\
E             & 99.8    & 52.8 \footnotesize{± 9.96} & 53.89 \footnotesize{± 5.39} & 49.34 ± 11.60 & 97.4 \footnotesize{± 2.0}    & 76.83 \footnotesize{± 2.32} \\
F             & 99.9    &             & 95.74 \footnotesize{± 0.75} & 94.16 \footnotesize{± 1.25} & 99.1 \footnotesize{± 0.6}     & 98.67 \footnotesize{± 0.05} \\
H             & 22.9    &             & 21.95 \footnotesize{± 0.03} & 76.84 ± 26.94 & 98.4 \footnotesize{± 1.1}    & 20.98 \footnotesize{± 1.98} \\
\bottomrule
\end{tabular}
}
\caption{Additional related work comparisons. Splits G and I are not included.}
\label{tab:additional_results}
\end{table*}

In this section of the appendix, we describe in more detail other related 
work on gSCAN and provide the results reported by those works
in Table \ref{tab:additional_results} for easier comparison
with our experimental results.

\paragraph{Modular} A recent work by \citet{conf/cogsci/Ruis22}. It uses a specialized decomposition into
Perception, Interaction, Navigation and Transformation Modules, each of which are trained independently with their
own training outputs, then connected together at test time. The modular decomposition gives a prior on how the problem
should be solved (for example by decomposition into egocentric and allocentric plans). The work also describes how
data augmentation can be used to improve the model, but we show the results coming from use of the modular architecture
alone. This approach can get good performance on Splits G and H. Performance on other splits is either slightly improved or comparable to the baseline in \citet{conf/nips/RuisABBL20}, which is likely due to the use of a similar underlying architecture of RNNs and CNNs as feature encoders.

\paragraph{Role-Guided} \citep{conf/emnlp/KuoKB21} This approach uses linguistic priors to decompose the parsing problem and specify
how sub-parsers are connected. It can achieve some level of performance on Split D and comparable performance on
Split H to the Transformer.

\paragraph{ViLBERT} is an adaptation of the ViLBERT model
for gSCAN by \citet{conf/emnlp/QiuH0SS21} and extended on by \citet{conf/emnlp/Sikarwar22}.
The state is first one-hot encoded, a few 2D convolution
layers are applied to it. The state is then flattened
and the channel values for each pixel are treated as
vectors for each location in the state. Afterwards,
there are several layers of cross-attention between the
instruction tokens and the state tokens. The cross-attented
representations are concatenated together and used as
input to a causal Transformer decoder to decode the
outputs.

\paragraph{GECA} Also known as ``Good Enough Compositional Augmentation"
(\citet{conf/acl/Andreas20}), applied to gSCAN by
\citet{conf/nips/RuisABBL20}. GECA is an augmentation method
which recognizes \textit{template fragments} in text, then realizes those
templates with other possible substitutions. Following the example in that work, if a dataset
contains ``she \underline{picks} the wug \underline{up} in Fresno``
and ``she \underline{puts} the wug \underline{down}
in Tempe", then the augmentation method generates samples of \underline{puts down}
substituted into sentences containing \underline{picks up}. For example
the sentence ``Pat \underline{picks} cats \underline{up}" can be augmented to
``Pat \underline{puts} cats \underline{down}". GECA relies on being able to
identify templates containing discontiguous fragments which contain at least two tokens.
In the case of SCAN, GECA might identify the fragment ``jump ... \texttt{JUMP ... JUMP ... JUMP}"
from the concatenated instruction-action pair ``jump thrice \texttt{JUMP JUMP JUMP}" and substitute it into ``walk around right thrice \texttt{WALK RTURN WALK RTURN WALK RTURN}" such that it is augmented into ``\underline{jump} around right thrice
\texttt{\underline{JUMP} RTURN \underline{JUMP} RTURN \underline{JUMP} RTURN}". As noted by
\citet{conf/acl/Andreas20}, the time and space complexity of GECA can be quite large
and scales with the number of recognized templates and fragments. The results reported by \citet{conf/nips/RuisABBL20} 
when using GECA in Table \ref{tab:additional_results} are
possibly out of date, since they were generated using
an RNN architecture as opposed to a Transformer, where better
performance on Splits B, C, E and F has been observed.
Also, GECA was only applied to the instructions and state and 
not to the target commands. Possibly the reason for this
is that the computational and memory complexity of GECA makes
it difficult to apply the joint space of the state, instruction
and target commands as in gSCAN.

{\color{black}\paragraph{CMA-ES} uses sparse hard attention with CMA-ES as the optimization algorithm
as opposed to a gradient-based optimizer. The model architecture consists only
of a multi-layer perceptron, predicting the next token with attention over the generated
output sequence. The method requires some supervision on what
the goal object is, in contrast with other approaches. Its strengths are that
convergence can happen very quickly and optimization can be run on lighter hardware.
The method also gets very good performance on Split H, however, as of the time of
writing, the authors have not yet published their code and did not provide any
analysis in their paper as to why the measured
Split H performance was so good, so it is not possible to make a detailed comparison with
our work.}

\begin{table*}[ht]
\resizebox{\textwidth}{!}{
\begin{tabular}{lcccccc}
              & ViLBERT          & Modular       & Role-guided & Transformer (ours)       & DemoGen     \\ 
%\hline
 & \citep{conf/emnlp/QiuH0SS21}  & \citep{conf/cogsci/Ruis22} & \citep{conf/emnlp/KuoKB21} & Ours  & Ours   \\ 
\hline
Learning Rate & 0.0015       & 0.001         & 0.001       & 0.0001          & 0.0001                    \\
Embedding Dim & 128          & 128           & 128         & 512             & 512 \\
Dropout & 0.1       & -         & -       & 0.1          & 0.1                    \\
Batch Size    & 128          & 200           & 200         & 128            & 128                        \\
Steps         & 114.96K      & 73K          & 150K        & 300K              & 300K                         \\
\#params      & 3M           &               &             & 88.3M           & 88.3M                       \\ 
\end{tabular}
}
\caption{Hyperparameters used in our experiments and the related work}
\label{tab:hyperparameters}
\end{table*}

\section{Experimental Details}
\label{sec:experimental_details}

We ran experiments to determine the performance of our approach. The Transformer blocks use an
embedding size ($d_{\textrm{model}}$) of 512 units and fully-connected layer
size ($d_{\textrm{FF}}$) of 2048 units is used. We use 12 layers for each of the encoder and decoder of the encoder-decoder transformer.
The learning rate is $10^{-5}$, we have an effective batch size of 128,
and training iteration count of 300,000. During training, dropout is not used
and weight decay is set to $10^{-3}$ with the AdamW optimizer.
Beta values are left at their defaults, $\beta_1 = 0.9$
and $\beta_2 = 0.999$.
Learning rate warmup is used up to step 30,000 to a peak learning rate
of $10^{-5}$, then decayed on a log-linear schedule from steps 30,000 to
300,000 to $10^{-6}$.  Gradient norms are clipped at
0.2 to improve training stability. We use 16-bit precision during 
training and make use of gradient accumulation in order
to simulate large batch sizes where memory is limited.

\section{Implementation of GandR for gSCAN}
\label{appendix:gandr}
We make small adaptations to GandR \cite{conf/coling/Zemlyanskiy22} to adapt it to the grounded setting. The baseline transformer model makes an initial prediction
for the query input, then the query input and prediction are vector-encoded and used to find other similar query-output pairs using the
index, which become the support inputs and outputs used for ICL. Compared to the original, we keep the $\alpha$ trade-off between input and target components fixed as opposed to varying it. We also don't include the state in the vector though the identity of the target
object and also its distance to the agent will likely be similar
as we select on the basis of input and output similarity. There is also
nothing to ensure that a diversity of different instructions is sampled -
only the near neighbours are sampled, even if they all correspond to a
single instruction.

\section{Implementation of Set-BSR (CovR) for gSCAN}
\label{appendix:covr}
We implement the main idea behind Set-BSR \cite{journals/corr/abs-2305-14907}
for the grounded setting. States are vector-encoded
and projected using PCA into 320 dimensions.
Instructions are TF-IDF encoded into vectors. Both
are concatenated with each other to make a vector
representation of an example. The instruction
component of the vector is weighted with $\alpha = 0.125$. The training-set vectors are placed into an
inner-product index. For performance reasons, we use a
Voronoi index with 512 cells and 10 cell probes per
search. For each vector in a split, we search
the index for the 128 nearest neighbours, sort
the neighbours in descending order according to
the number of matching two-grams, one-grams and
the cosine similarity to the query state. Then we
pick the top $k = 16$ examples as the support set.

\vfill\null
\pagebreak

\vfill\null
\pagebreak

\section{Properties of Generated Demonstrations, other splits}
\label{sec:generated_demonstrations_other_splits}

Properties of Generated Demonstrations for the other splits are shown in
tables below.

\begin{table}[H]
\resizebox{\columnwidth}{!}{
\begin{tabular}{lrrrrrr}
\toprule
\multicolumn{6}{c}{Split A} \\
\toprule
 & DemoG & GandR & CovR & Expert & OtherS & Random \\
\midrule
\footnotesize{(1) Desc. Obj.} & \footnotesize{0.32} & \footnotesize{0.83} & \footnotesize{0.15} & \footnotesize{1.00} & \footnotesize{1.00} & \footnotesize{0.07} \\
\footnotesize{(2) Agent Pos.} & \footnotesize{1.00} & \footnotesize{0.07} & \footnotesize{1.00} & \footnotesize{1.00} & \footnotesize{0.03} & \footnotesize{1.00} \\
\footnotesize{(3) Tgt. Pos.} & \footnotesize{0.37} & \footnotesize{0.08} & \footnotesize{0.27} & \footnotesize{1.00} & \footnotesize{0.03} & \footnotesize{0.07} \\
\footnotesize{(4) Same Diff.} & \footnotesize{0.37} & \footnotesize{0.31} & \footnotesize{0.27} & \footnotesize{1.00} & \footnotesize{0.02} & \footnotesize{0.07} \\
\footnotesize{(5) Tgt. Obj.} & \footnotesize{0.37} & \footnotesize{0.26} & \footnotesize{0.22} & \footnotesize{1.00} & \footnotesize{0.25} & \footnotesize{0.07} \\
\footnotesize{(6) Verb \& (5)} & \footnotesize{1.00} & \footnotesize{0.93} & \footnotesize{0.91} & \footnotesize{1.00} & \footnotesize{0.50} & \footnotesize{0.07} \\
\footnotesize{(7) Advb \& (5)} & \footnotesize{0.75} & \footnotesize{0.93} & \footnotesize{0.77} & \footnotesize{1.00} & \footnotesize{0.38} & \footnotesize{0.07} \\
\footnotesize{(8) (6) \& (7)} & \footnotesize{0.75} & \footnotesize{0.93} & \footnotesize{0.73} & \footnotesize{1.00} & \footnotesize{0.23} & \footnotesize{0.07} \\
\footnotesize{(9) (4) \& (8)} & \footnotesize{0.75} & \footnotesize{0.57} & \footnotesize{0.65} & \footnotesize{1.00} & \footnotesize{0.00} & \footnotesize{0.07} \\
\bottomrule
\end{tabular}
}
\end{table}

\begin{table}[H]
\resizebox{\columnwidth}{!}{
\begin{tabular}{lrrrrrr}
\toprule
\multicolumn{6}{c}{Split B} \\
\midrule
 & DemoG & GandR & CovR & Expert & OtherS & Random \\
\midrule
\footnotesize{(1) Desc. Obj.} & \footnotesize{0.26} & \footnotesize{0.00} & \footnotesize{0.00} & \footnotesize{1.00} & \footnotesize{0.00} & \footnotesize{0.00} \\
\footnotesize{(2) Agent Pos.} & \footnotesize{1.00} & \footnotesize{0.13} & \footnotesize{1.00} & \footnotesize{1.00} & \footnotesize{0.00} & \footnotesize{1.00} \\
\footnotesize{(3) Tgt. Pos.} & \footnotesize{0.32} & \footnotesize{0.15} & \footnotesize{0.29} & \footnotesize{1.00} & \footnotesize{0.00} & \footnotesize{0.00} \\
\footnotesize{(4) Same Diff.} & \footnotesize{0.32} & \footnotesize{0.44} & \footnotesize{0.29} & \footnotesize{1.00} & \footnotesize{0.00} & \footnotesize{0.00} \\
\footnotesize{(5) Tgt. Obj.} & \footnotesize{0.32} & \footnotesize{0.03} & \footnotesize{0.18} & \footnotesize{1.00} & \footnotesize{0.00} & \footnotesize{0.00} \\
\footnotesize{(6) Verb \& (5)} & \footnotesize{1.00} & \footnotesize{0.30} & \footnotesize{0.85} & \footnotesize{1.00} & \footnotesize{0.00} & \footnotesize{0.00} \\
\footnotesize{(7) Advb \& (5)} & \footnotesize{0.66} & \footnotesize{0.30} & \footnotesize{0.71} & \footnotesize{1.00} & \footnotesize{0.00} & \footnotesize{0.00} \\
\footnotesize{(8) (6) \& (7)} & \footnotesize{0.66} & \footnotesize{0.30} & \footnotesize{0.69} & \footnotesize{1.00} & \footnotesize{0.00} & \footnotesize{0.00} \\
\footnotesize{(9) (4) \& (8)} & \footnotesize{0.66} & \footnotesize{0.24} & \footnotesize{0.63} & \footnotesize{1.00} & \footnotesize{0.00} & \footnotesize{0.00} \\
\bottomrule
\end{tabular}
}
\end{table}

\begin{table}[H]
\resizebox{\columnwidth}{!}{
\begin{tabular}{lrrrrrr}
\toprule
\multicolumn{6}{c}{Split C} \\
\midrule
 & DemoG & GandR & CovR & Expert & OtherS & Random \\
\midrule
\footnotesize{(1) Desc. Obj.} & \footnotesize{0.16} & \footnotesize{0.47} & \footnotesize{0.15} & \footnotesize{1.00} & \footnotesize{1.00} & \footnotesize{0.15} \\
\footnotesize{(2) Agent Pos.} & \footnotesize{1.00} & \footnotesize{0.12} & \footnotesize{1.00} & \footnotesize{1.00} & \footnotesize{0.03} & \footnotesize{1.00} \\
\footnotesize{(3) Tgt. Pos.} & \footnotesize{0.19} & \footnotesize{0.13} & \footnotesize{0.18} & \footnotesize{1.00} & \footnotesize{0.03} & \footnotesize{0.15} \\
\footnotesize{(4) Same Diff.} & \footnotesize{0.19} & \footnotesize{0.44} & \footnotesize{0.18} & \footnotesize{1.00} & \footnotesize{0.02} & \footnotesize{0.15} \\
\footnotesize{(5) Tgt. Obj.} & \footnotesize{0.19} & \footnotesize{0.00} & \footnotesize{0.00} & \footnotesize{1.00} & \footnotesize{0.00} & \footnotesize{0.15} \\
\footnotesize{(6) Verb \& (5)} & \footnotesize{0.79} & \footnotesize{0.00} & \footnotesize{0.00} & \footnotesize{1.00} & \footnotesize{0.00} & \footnotesize{0.15} \\
\footnotesize{(7) Advb \& (5)} & \footnotesize{0.41} & \footnotesize{0.00} & \footnotesize{0.00} & \footnotesize{1.00} & \footnotesize{0.00} & \footnotesize{0.15} \\
\footnotesize{(8) (6) \& (7)} & \footnotesize{0.40} & \footnotesize{0.00} & \footnotesize{0.00} & \footnotesize{1.00} & \footnotesize{0.00} & \footnotesize{0.15} \\
\footnotesize{(9) (4) \& (8)} & \footnotesize{0.40} & \footnotesize{0.00} & \footnotesize{0.00} & \footnotesize{1.00} & \footnotesize{0.00} & \footnotesize{0.15} \\
\bottomrule
\end{tabular}
}
\end{table}

\begin{table}[H]
\resizebox{\columnwidth}{!}{
\begin{tabular}{lrrrrrr}
\toprule
\multicolumn{6}{c}{Split D} \\
\midrule
 & DemoG & GandR & CovR & Expert & OtherS & Random \\
\midrule
\footnotesize{(1) Desc. Obj.} & \footnotesize{0.19} & \footnotesize{0.83} & \footnotesize{0.18} & \footnotesize{1.00} & \footnotesize{1.00} & \footnotesize{0.16} \\
\footnotesize{(2) Agent Pos.} & \footnotesize{1.00} & \footnotesize{0.03} & \footnotesize{1.00} & \footnotesize{1.00} & \footnotesize{0.02} & \footnotesize{1.00} \\
\footnotesize{(3) Tgt. Pos.} & \footnotesize{0.33} & \footnotesize{0.03} & \footnotesize{0.00} & \footnotesize{1.00} & \footnotesize{0.02} & \footnotesize{0.16} \\
\footnotesize{(4) Same Diff.} & \footnotesize{0.33} & \footnotesize{0.00} & \footnotesize{0.00} & \footnotesize{1.00} & \footnotesize{0.00} & \footnotesize{0.16} \\
\footnotesize{(5) Tgt. Obj.} & \footnotesize{0.33} & \footnotesize{0.20} & \footnotesize{0.05} & \footnotesize{1.00} & \footnotesize{0.10} & \footnotesize{0.16} \\
\footnotesize{(6) Verb \& (5)} & \footnotesize{0.99} & \footnotesize{0.89} & \footnotesize{0.42} & \footnotesize{1.00} & \footnotesize{0.25} & \footnotesize{0.16} \\
\footnotesize{(7) Advb \& (5)} & \footnotesize{0.89} & \footnotesize{0.88} & \footnotesize{0.25} & \footnotesize{1.00} & \footnotesize{0.17} & \footnotesize{0.16} \\
\footnotesize{(8) (6) \& (7)} & \footnotesize{0.89} & \footnotesize{0.88} & \footnotesize{0.20} & \footnotesize{1.00} & \footnotesize{0.06} & \footnotesize{0.16} \\
\footnotesize{(9) (4) \& (8)} & \footnotesize{0.89} & \footnotesize{0.00} & \footnotesize{0.00} & \footnotesize{1.00} & \footnotesize{0.00} & \footnotesize{0.16} \\
\bottomrule
\end{tabular}
}
\end{table}

\begin{table}[H]
\resizebox{\columnwidth}{!}{
\begin{tabular}{lrrrrrr}
\toprule
\multicolumn{6}{c}{Split E} \\
\midrule
 & DemoG & GandR & CovR & Expert & OtherS & Random \\
\midrule
\footnotesize{(1) Desc. Obj.} & \footnotesize{0.22} & \footnotesize{0.89} & \footnotesize{0.07} & \footnotesize{1.00} & \footnotesize{0.00} & \footnotesize{0.00} \\
\footnotesize{(2) Agent Pos.} & \footnotesize{1.00} & \footnotesize{0.11} & \footnotesize{1.00} & \footnotesize{1.00} & \footnotesize{0.00} & \footnotesize{1.00} \\
\footnotesize{(3) Tgt. Pos.} & \footnotesize{0.27} & \footnotesize{0.12} & \footnotesize{0.22} & \footnotesize{1.00} & \footnotesize{0.00} & \footnotesize{0.00} \\
\footnotesize{(4) Same Diff.} & \footnotesize{0.27} & \footnotesize{0.35} & \footnotesize{0.22} & \footnotesize{1.00} & \footnotesize{0.00} & \footnotesize{0.00} \\
\footnotesize{(5) Tgt. Obj.} & \footnotesize{0.27} & \footnotesize{0.03} & \footnotesize{0.14} & \footnotesize{1.00} & \footnotesize{0.00} & \footnotesize{0.00} \\
\footnotesize{(6) Verb \& (5)} & \footnotesize{0.96} & \footnotesize{0.20} & \footnotesize{0.81} & \footnotesize{1.00} & \footnotesize{0.00} & \footnotesize{0.00} \\
\footnotesize{(7) Advb \& (5)} & \footnotesize{0.50} & \footnotesize{0.20} & \footnotesize{0.63} & \footnotesize{1.00} & \footnotesize{0.00} & \footnotesize{0.00} \\
\footnotesize{(8) (6) \& (7)} & \footnotesize{0.50} & \footnotesize{0.20} & \footnotesize{0.60} & \footnotesize{1.00} & \footnotesize{0.00} & \footnotesize{0.00} \\
\footnotesize{(9) (4) \& (8)} & \footnotesize{0.50} & \footnotesize{0.14} & \footnotesize{0.50} & \footnotesize{1.00} & \footnotesize{0.00} & \footnotesize{0.00} \\
\bottomrule
\end{tabular}
}
\end{table}

\begin{table}[H]
\resizebox{\columnwidth}{!}{
\begin{tabular}{lrrrrrr}
\toprule
\multicolumn{6}{c}{Split F} \\
\midrule
 & DemoG & GandR & CovR & Expert & OtherS & Random \\
\midrule
\footnotesize{(1) Desc. Obj.} & \footnotesize{0.26} & \footnotesize{0.81} & \footnotesize{0.23} & \footnotesize{1.00} & \footnotesize{1.00} & \footnotesize{0.15} \\
\footnotesize{(2) Agent Pos.} & \footnotesize{1.00} & \footnotesize{0.12} & \footnotesize{1.00} & \footnotesize{1.00} & \footnotesize{0.03} & \footnotesize{1.00} \\
\footnotesize{(3) Tgt. Pos.} & \footnotesize{0.33} & \footnotesize{0.15} & \footnotesize{0.26} & \footnotesize{1.00} & \footnotesize{0.03} & \footnotesize{0.15} \\
\footnotesize{(4) Same Diff.} & \footnotesize{0.33} & \footnotesize{0.37} & \footnotesize{0.26} & \footnotesize{1.00} & \footnotesize{0.02} & \footnotesize{0.15} \\
\footnotesize{(5) Tgt. Obj.} & \footnotesize{0.33} & \footnotesize{0.00} & \footnotesize{0.10} & \footnotesize{1.00} & \footnotesize{0.07} & \footnotesize{0.15} \\
\footnotesize{(6) Verb \& (5)} & \footnotesize{0.96} & \footnotesize{0.00} & \footnotesize{0.00} & \footnotesize{1.00} & \footnotesize{0.00} & \footnotesize{0.15} \\
\footnotesize{(7) Advb \& (5)} & \footnotesize{0.60} & \footnotesize{0.00} & \footnotesize{0.62} & \footnotesize{1.00} & \footnotesize{0.29} & \footnotesize{0.15} \\
\footnotesize{(8) (6) \& (7)} & \footnotesize{0.58} & \footnotesize{0.00} & \footnotesize{0.00} & \footnotesize{1.00} & \footnotesize{0.00} & \footnotesize{0.15} \\
\footnotesize{(9) (4) \& (8)} & \footnotesize{0.58} & \footnotesize{0.00} & \footnotesize{0.00} & \footnotesize{1.00} & \footnotesize{0.00} & \footnotesize{0.15} \\
\bottomrule
\end{tabular}
}
\end{table}

\begin{table}[H]
\resizebox{\columnwidth}{!}{
\begin{tabular}{lrrrrrr}
\toprule
\multicolumn{6}{c}{Split G} \\
\midrule
 & DemoG & GandR & CovR & Expert & OtherS & Random \\
\midrule
\footnotesize{(1) Desc. Obj.} & \footnotesize{0.39} & \footnotesize{0.91} & \footnotesize{0.31} & \footnotesize{1.00} & \footnotesize{1.00} & \footnotesize{0.20} \\
\footnotesize{(2) Agent Pos.} & \footnotesize{1.00} & \footnotesize{0.14} & \footnotesize{1.00} & \footnotesize{1.00} & \footnotesize{0.03} & \footnotesize{1.00} \\
\footnotesize{(3) Tgt. Pos.} & \footnotesize{0.50} & \footnotesize{0.16} & \footnotesize{0.37} & \footnotesize{1.00} & \footnotesize{0.03} & \footnotesize{0.20} \\
\footnotesize{(4) Same Diff.} & \footnotesize{0.50} & \footnotesize{0.35} & \footnotesize{0.37} & \footnotesize{1.00} & \footnotesize{0.02} & \footnotesize{0.20} \\
\footnotesize{(5) Tgt. Obj.} & \footnotesize{0.50} & \footnotesize{0.22} & \footnotesize{0.24} & \footnotesize{1.00} & \footnotesize{0.20} & \footnotesize{0.20} \\
\footnotesize{(6) Verb \& (5)} & \footnotesize{1.00} & \footnotesize{0.91} & \footnotesize{0.93} & \footnotesize{1.00} & \footnotesize{0.51} & \footnotesize{0.20} \\
\footnotesize{(7) Advb \& (5)} & \footnotesize{0.00} & \footnotesize{0.01} & \footnotesize{0.00} & \footnotesize{1.00} & \footnotesize{0.00} & \footnotesize{0.20} \\
\footnotesize{(8) (6) \& (7)} & \footnotesize{0.00} & \footnotesize{0.01} & \footnotesize{0.00} & \footnotesize{1.00} & \footnotesize{0.00} & \footnotesize{0.20} \\
\footnotesize{(9) (4) \& (8)} & \footnotesize{0.00} & \footnotesize{0.00} & \footnotesize{0.00} & \footnotesize{1.00} & \footnotesize{0.00} & \footnotesize{0.20} \\
\bottomrule
\end{tabular}
}
\end{table}

\begin{table}[H]
\resizebox{\columnwidth}{!}{
\begin{tabular}{lrrrrrr}
\toprule
\multicolumn{6}{c}{Split H} \\
\midrule
 & DemoG & GandR & CovR & Expert & OtherS & Random \\
\midrule
\footnotesize{(1) Desc. Obj.} & \footnotesize{0.33} & \footnotesize{0.68} & \footnotesize{0.33} & \footnotesize{1.00} & \footnotesize{1.00} & \footnotesize{0.16} \\
\footnotesize{(2) Agent Pos.} & \footnotesize{1.00} & \footnotesize{0.08} & \footnotesize{1.00} & \footnotesize{1.00} & \footnotesize{0.03} & \footnotesize{1.00} \\
\footnotesize{(3) Tgt. Pos.} & \footnotesize{0.44} & \footnotesize{0.08} & \footnotesize{0.39} & \footnotesize{1.00} & \footnotesize{0.03} & \footnotesize{0.16} \\
\footnotesize{(4) Same Diff.} & \footnotesize{0.44} & \footnotesize{0.09} & \footnotesize{0.39} & \footnotesize{1.00} & \footnotesize{0.02} & \footnotesize{0.16} \\
\footnotesize{(5) Tgt. Obj.} & \footnotesize{0.44} & \footnotesize{0.14} & \footnotesize{0.27} & \footnotesize{1.00} & \footnotesize{0.19} & \footnotesize{0.16} \\
\footnotesize{(6) Verb \& (5)} & \footnotesize{1.00} & \footnotesize{0.15} & \footnotesize{0.88} & \footnotesize{1.00} & \footnotesize{0.43} & \footnotesize{0.16} \\
\footnotesize{(7) Advb \& (5)} & \footnotesize{0.88} & \footnotesize{0.51} & \footnotesize{0.78} & \footnotesize{1.00} & \footnotesize{0.33} & \footnotesize{0.16} \\
\footnotesize{(8) (6) \& (7)} & \footnotesize{0.88} & \footnotesize{0.00} & \footnotesize{0.70} & \footnotesize{1.00} & \footnotesize{0.19} & \footnotesize{0.16} \\
\footnotesize{(9) (4) \& (8)} & \footnotesize{0.88} & \footnotesize{0.00} & \footnotesize{0.62} & \footnotesize{1.00} & \footnotesize{0.00} & \footnotesize{0.16} \\
\bottomrule
\end{tabular}
}
\end{table}

\vfill\null
\pagebreak

\section{Heuristic Function}
\label{sec:appendix_oracle}

The \textbf{Heuristic} function generates relevant instructions by the use of a templating mechanism, which replaces
verbs and adverbs in the sentence with other verbs and adverbs, such that the whole combination is still
in distribution, but not the same as the query instruction. The rules of the system are:
\begin{itemize}
    \item Replace ``pull" with ``push" and ``walk to"
    \item Replace ``walk to" with ``push" and ``pull" (but not if ``while spinning" is the adverb)
    \item Replace ``push" with ``walk to" and ``pull" (but not if ``while spinning" is the adverb)
    \item Replace ``while zigzagging" with ``hesitantly", nothing and ``while spinning" (but not if ``push" is the verb)
    \item Replace ``hesitantly" with ``while zigzagging", nothing and ``while spinning" (but not if ``push" is the verb)
    \item Replace ``while spinning" with ``hesitantly", ``while zigzagging" and nothing
\end{itemize}

Examples of what the oracle function generates for a given query instruction and environment can be found
in Figure \ref{fig:generation_examples}. Actions are generated by using the same procedure provided in \citet{conf/nips/RuisABBL20}. \textcolor{black}{The instruction generated by the oracle is given to the demonstration generation procedure and a demonstration is generated by that. A demonstration can also be generated by providing the oracle-generated instruction and current state representation as input to a Transformer model trained on the provided training set}.

\section{Permuter Blocks}
\label{sec:appendix_permuter}

\begin{table}[t]
\centering
\resizebox{\linewidth}{!}{
\begin{tabular}{cc|cc}
Word & Symbol & Action & Symbol \\
\hline
`a' & 0 & \texttt{PULL} & 0 \\
`big' & 1 & \texttt{PUSH} & 1 \\
`blue' & 2 & \texttt{STAY} & 2 \\
`cautiously' & 3 & \texttt{LTURN} & 3 \\
`circle' & 4 & \texttt{RTURN} & 4 \\
`cylinder` & 5 & \texttt{WALK} & 5 \\
`green' & 6 & &  \\
`hesitantly' & 7 & & \\
`pull' & 8 & & \\
`push & 9  & & \\
`red' & 10 & & \\
`small' & 11  & & \\
`square' & 12  & & \\
`to' & 13  & & \\
`walk' & 14  & & \\
`while spinning' & 15 & & \\
`while zigzagging` & 16  & & \\
\end{tabular}
}
\caption{\label{tab:dictionary}Default mapping of words and actions to symbols}
\end{table}

\begin{table*}[ht]
\centering
\resizebox{\textwidth}{!}{
\begin{tabular}{|p{5cm}|p{7cm}|p{3cm}|p{3cm}|}
\hline
\multicolumn{1}{|c|}{Original actions} & \multicolumn{1}{|c|}{Permutation} & \multicolumn{1}{|c|}{Encoded actions} & \multicolumn{1}{|c|}{Permuted encoding} \\
\hline
WALK(5) RTURN WALK(5) & $\text{PULL(0)} \to 0, $ $\text{PUSH(1)} \to 5, $ $\text{STAY(2)} \to 2, $ $\text{LTURN(3)} \to 1, $ $\text{RTURN(4)} \to 3, $ $\text{WALK(5)} \to 4, $ & 5(5) 4 5(5) & 4(5) 3 4(5) \\ \hline
RTURN WALK(3) & $\text{PULL(0)} \to 0, $ $\text{PUSH(1)} \to 2, $ $\text{STAY(2)} \to 3, $ $\text{LTURN(3)} \to 5, $ $\text{RTURN(4)} \to 4, $ $\text{WALK(5)} \to 1, $ & 4 5(3) & 4 1(3) \\ \hline
LTURN(4) WALK LTURN(4) WALK LTURN(5) WALK LTURN(4) WALK LTURN(4) WALK LTURN(4) WALK LTURN(4) WALK & $\text{PULL(0)} \to 4, $ $\text{PUSH(1)} \to 5, $ $\text{STAY(2)} \to 0, $ $\text{LTURN(3)} \to 2, $ $\text{RTURN(4)} \to 3, $ $\text{WALK(5)} \to 1, $ & 3(4) 5 3(4) 5 3(5) 5 3(4) 5 3(4) 5 3(4) 5 3(4) 5 & 2(4) 1 2(4) 1 2(5) 1 2(4) 1 2(4) 1 2(4) 1 2(4) 1 \\ \hline
LTURN WALK STAY WALK STAY WALK STAY WALK STAY & $\text{PULL(0)} \to 3, $ $\text{PUSH(1)} \to 0, $ $\text{STAY(2)} \to 2, $ $\text{LTURN(3)} \to 5, $ $\text{RTURN(4)} \to 4, $ $\text{WALK(5)} \to 1, $ & 3 5 2 5 2 5 2 5 2 & 5 1 2 1 2 1 2 1 2 \\ \hline
LTURN WALK STAY WALK STAY & $\text{PULL(0)} \to 0, $ $\text{PUSH(1)} \to 3, $ $\text{STAY(2)} \to 4, $ $\text{LTURN(3)} \to 5, $ $\text{RTURN(4)} \to 2, $ $\text{WALK(5)} \to 1, $ & 3 5 2 5 2 & 5 1 4 1 4 \\ \hline
LTURN(4) WALK LTURN(4) WALK LTURN(4) WALK LTURN(4) RTURN WALK LTURN(4) WALK LTURN(4) WALK LTURN(4) WALK LTURN(4) WALK & $\text{PULL(0)} \to 0, $ $\text{PUSH(1)} \to 4, $ $\text{STAY(2)} \to 5, $ $\text{LTURN(3)} \to 1, $ $\text{RTURN(4)} \to 3, $ $\text{WALK(5)} \to 2, $ & 3(4) 5 3(4) 5 3(4) 5 3(4) 4 5 3(4) 5 3(4) 5 3(4) 5 3(4) 5 & 1(4) 2 1(4) 2 1(4) 2 1(4) 3 2 1(4) 2 1(4) 2 1(4) 2 1(4) 2 \\ \hline
LTURN WALK(2) PUSH & $\text{PULL(0)} \to 1, $ $\text{PUSH(1)} \to 0, $ $\text{STAY(2)} \to 5, $ $\text{LTURN(3)} \to 3, $ $\text{RTURN(4)} \to 4, $ $\text{WALK(5)} \to 2, $ & 3 5(2) 1 & 3 2(2) 0 \\ \hline
\end{tabular}
}
\caption{\label{tab:permuter_block_actions} Actions and possible mapping permutations generated by the permuter block.}
\end{table*}

The permuter block shuffles the indices mapping words to symbols in the dictionary given in Table \ref{tab:dictionary}. Table \ref{tab:permuter_block_actions} gives an example of how the permuted sequences might look to the encoders. Essentially the individual symbols no longer hold any special meaning without reference to the demonstrations, only conditional autoregressive probabilities up to a permutation hold meaning.

\section{Natural-ish Language gSCAN Dataset}
\label{sec:appendix_gen_data}
The dataset is generated by extracting all of the input sentences from
gSCAN and its derivatives, then using the commercial \texttt{gpt3.5-turbo}
model from OpenAI\footnote{As of 5 May 2023} to generate additional paraphrases
of the input sentence. The paraphrases are generated by creating four dataset
specific prompts, each with an 10 examples of how one instruction in the dataset
may be paraphrased, then requesting 25 additional paraphrases for a different
instruction in the same dataset to be completed by
the language model. The prompts are given in Appendix \ref{appendix:prompts}. The
prompts modes are described as follows:

\paragraph{Simple} Paraphrases of ``Push a red square"
\paragraph{Adverb} Paraphrases of ``Push a red square cautiously"
\paragraph{Relational} Paraphrases of ``Push a red circle that is south east of a blue circle"
\paragraph{ReaSCAN} Paraphrases of ``Pull the yellow square that is inside of a big red box and in the same row as a small red circle and in the same column as a small cylinder while spinning"

The 10 paraphrase examples were written by ourselves - the idea is that they
show how adverbs and actions can be replaced by synonyms, and also show
examples of the same instruction in a different sentence ordering. For example,
``push a red square" can be paraphrased as ``shove the red square" or ``Walk to a red 
square and push it". The paraphrases can also include additional verbs adverbs which 
are distractors, for example ``grasp a red square and move it along".

\begin{table}[H]
\centering
\begin{tabular}{l|ccc}
\hline
  & gSCAN & RS & ReaSCAN \\
\hline
Uniq. Instrs. & 430 & 31799 & 4381 \\
Uniq. Tmpls. & - & 21 & 658 \\
\hline
Gen. Instrs. & 12778 & 731377 & 99698 \\
Gen. Tmpls. & - & 483 & 14683 \\
\hline
Prompt & Simple & Relational & ReaSCAN \\
\hline
\end{tabular}
\caption{Generation properties and configuration for each of the datasets}
\label{tab:gen_props}
\end{table}

We generate paraphrases of instructions in gSCAN, gSCAN-RS and ReaSCAN. The default
generation mode creates paraphrases for each unique instruction individually.
However for gSCAN-RS and ReaSCAN, the number of unique instructions is very large,
which would mean that generation would come at both a high time and monetary cost.
The reason for this is the combinatorial explosion of factors; in gSCAN-RS the
instructions are given as target objects with positions relative to other objects.
To address this problem, we also have a "template" generation mode, which replaces
the object descriptions (size, color, shape) with a template placeholder, generates
paraphrases for the templates, then generates ``realised" paraphrases from those
templates for each of the corresponding object descriptions.  This reduces the
number of requests to the model from hundreds of thousands to thousands.

\section{Prompts used for GPT3.5}
\label{appendix:prompts}

\subsection{gSCAN Simple Prompt}

Here are 10 similar statements to “push a red square"

\begin{enumerate}
\item Push the red square
\item Move a red square
\item Shove the red square
\item Go to the red square and shove it
\item Go to the red square and push it
\item Walk to the red square and push it
\item Find a red square and push it
\item Locate a red square and push it
\item Get to the red square and move it along
\item Walk up to the red square and then really push it
\end{enumerate}

Can you generate 25 similar statements for “\{\{QUERY\}\}” in English?

\section{Examples of Generated Paraphrases}
\label{appendix:paraphrases}

Examples: 367933
Unique Instructions: 430
Templates: 828
Sample Responses:
\paragraph{Original} walk to a yellow small cylinder hesitantly
\begin{enumerate}
\item Walk hesitantly to a small yellow cylinder
\item Approach a small yellow cylinder with hesitation
\item Take cautious steps towards a small yellow cylinder
\item Walk slowly towards a small yellow cylinder
\item Move towards a small yellow cylinder with caution
\item Walk towards a small yellow cylinder carefully
\item Take hesitant steps towards a small yellow cylinder
\item Approach a small yellow cylinder slowly and hesitantly
\item Walk towards a small yellow cylinder with apprehension
\item Move towards a small yellow cylinder hesitantly
\item Walk to a small yellow cylinder with reluctance
\item Take tentative steps towards a small yellow cylinder
\item Walk towards a small yellow cylinder with uncertainty
\item Approach a small yellow cylinder with trepidation
\item Walk towards a small yellow cylinder with reservation
\item Take hesitant strides towards a small yellow cylinder
\item Walk to a small yellow cylinder with caution
\item Move towards a small yellow cylinder with unease
\item Walk towards a small yellow cylinder with doubt
\item Approach a small yellow cylinder with timidity
\item Walk towards a small yellow cylinder with hesitance
\item Take slow steps towards a small yellow cylinder
\item Walk towards a small yellow cylinder with wariness
\item Move towards a small yellow cylinder with hesitancy
\item Walk towards a small yellow cylinder with reluctance and caution.
\end{enumerate}

\paragraph{Original}  pull a circle hesitantly
\begin{enumerate}
\item Hesitantly pull a circle
\item Pull a circle with hesitation
\item Gently pull a circle
\item Pull a circle slowly and cautiously
\item Approach the circle and pull it hesitantly
\item Pull the circle with care
\item Pull the circle slowly and carefully
\item Take your time and pull the circle hesitantly
\item Pull the circle with a bit of hesitation
\item Approach the circle and pull it gently
\item Pull the circle with caution
\item Pull the circle carefully and hesitantly
\item Slowly pull the circle with hesitation
\item Pull the circle with a sense of caution
\item Pull the circle with a bit of reluctance
\item Pull the circle slowly and with care
\item Hesitate before pulling the circle
\item Pull the circle with a gentle touch
\item Pull the circle with a bit of apprehension
\item Pull the circle with a sense of uncertainty
\item Pull the circle with a bit of nervousness
\item Pull the circle with a bit of trepidation
\item Pull the circle with a sense of hesitation
\item Pull the circle with a bit of doubt
\item Pull the circle with a bit of reservation
\end{enumerate}

\paragraph{Original}  pull a circle while zigzagging
\begin{enumerate}
\item Zigzag while pulling a circle
\item Pull a circle in a zigzag pattern
\item Carefully pull a circle while zigzagging
\item Zigzag and pull a circle simultaneously
\item Pull a circle while moving in a zigzag motion
\item With caution, pull a circle while zigzagging
\item Zigzag your way to the circle and pull it
\item Pull a circle while making zigzag movements
\item Zigzag and pull the circle with care
\item Pull a circle while navigating in a zigzag direction
\item Move in a zigzag pattern while pulling a circle
\item Pull a circle while making a zigzag path
\item Zigzag towards the circle and pull it
\item Pull a circle while making zigzag turns
\item Carefully zigzag and pull the circle
\item Zigzag and carefully pull the circle
\item Pull a circle while making sharp zigzag movements
\item Zigzag and pull the circle with caution
\item Pull a circle while making quick zigzag motions
\item Zigzag and pull the circle slowly
\item Pull a circle while zigzagging in a controlled manner
\item Zigzag and pull the circle with precision
\item Pull a circle while making small zigzag movements
\item Zigzag and pull the circle with care and attention
\item Pull a circle while zigzagging smoothly.
\end{enumerate}

\section{Properties of Natural-ish Language gSCAN Dataset}
\label{sec:appendix_gen_data_props}

\subsection{Linguistic Properties}

In this section we examine the linguistic properties of the dataset. The main research question
is whether the instructions as paraphrased by GPT3.5 look more like natural language.
Clearly, the paraphrased data has greater vocabulary complexity. But merely substituting
words for synonyms would not make synthetic data appear any more natural, nor does it
pose any real challenges to a learning algorithm that would need to act on the instructions.
We examine two other indicia, unique parses and fit to a Zipf distribution of word frequency.

\begin{table}[H]
    \centering
    \begin{tabular}{lrrrr}
    \toprule
    {} &   parses &  words &  zipf a &  rmse \\
    \midrule
    \footnotesize{gSCAN}                &    18 &  18 &    1.99 &  0.11 \\
    \footnotesize{NL-gSCAN}    &  1550 & 859 &    1.29 &  0.01 \\
    \hline
    \footnotesize{SR}             &   234 &  20 &    1.90 &  0.10 \\
    \footnotesize{NL-SR} &  9785 & 126 &    1.40 &  0.03 \\
    \hline
    \footnotesize{ReaSCAN}              &  1400 &  35 &    1.26 &  0.04 \\
    \footnotesize{NL-ReaSCAN}  & 42759 & 631 &    1.22 &  0.01 \\
    \bottomrule
    \end{tabular}
    \caption{Linguistic properties of each dataset and its corresponding paraphrased (denoted NL-) dataset}.
    \label{tab:linguistic_properties}
\end{table}

\paragraph{Parses} We compute the number of unique parses among all the instructions in each
training set. A \textit{parse} is an assignment of word-role labels, indicating the
linguistic role of the token in the instruction. For example, a token may be an adjective,
an adverb or some sort of connector. The parses are computed over every instruction in the
training data using the spaCy package. As shown in Table~\ref{tab:linguistic_properties},
the number of unique parses in the paraphrased datasets are an order of magnitude larger
than the number of unique parses in the synthetic datasets. This reflects the diversity
of instruction structures that exist in the paraphrased datasets.

\begin{figure}[H]
\centering
\includegraphics[width=\columnwidth,keepaspectratio=1]{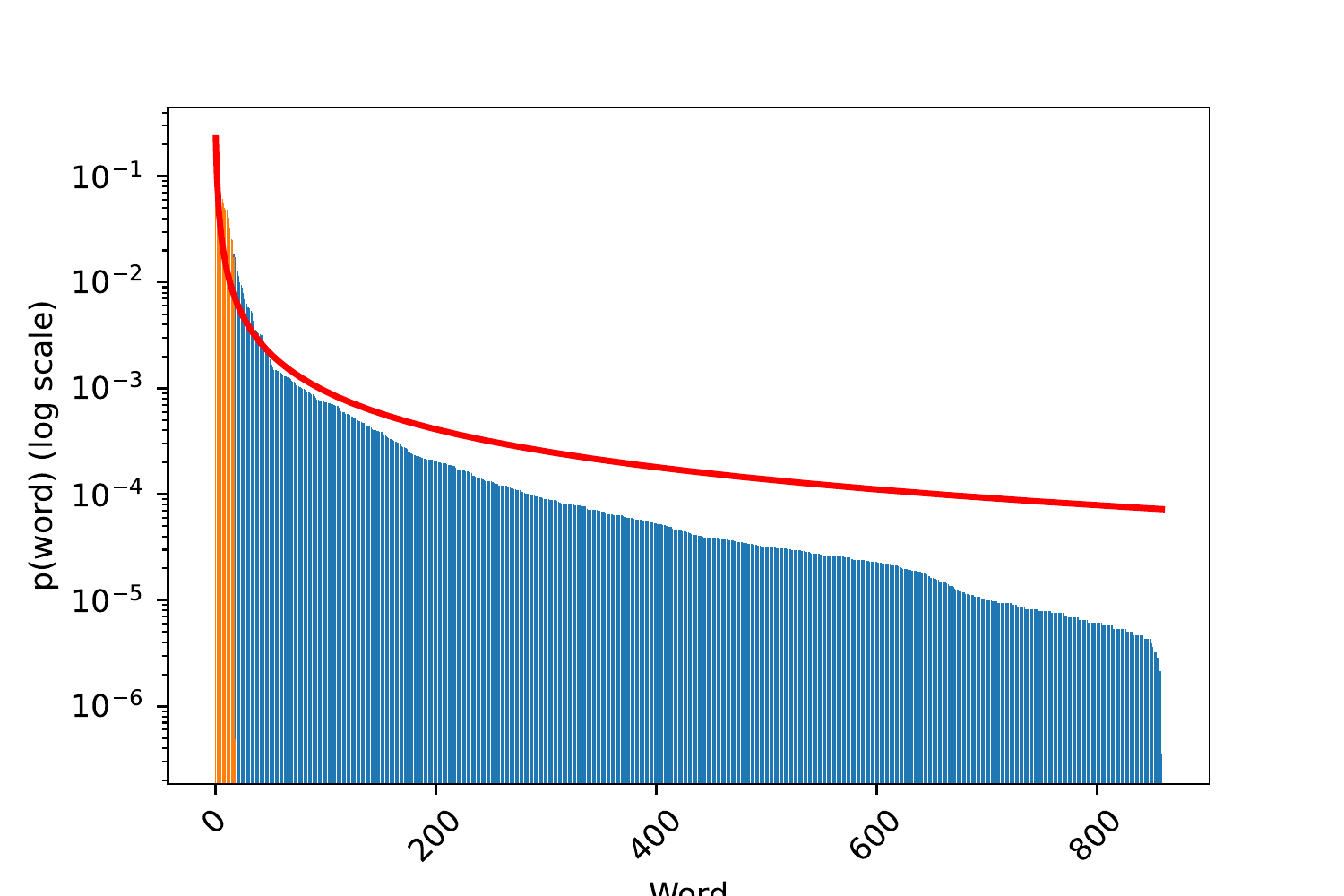}
\caption{Word frequency distribution of NL-gSCAN and gSCAN, each
compared to the best fitting Zipf distribution probability density function. gSCAN words are in orange and NL-gSCAN words are in blue (comprising of the larger vocabulary).}
\label{fig:gscan_zipf}
\end{figure}

\paragraph{Zipfian Distribution Fit} Natural language is hypothesized to fit a Zipfian
power-law distribution, where the probability of drawing a word from a corpus is inversely
proportional to its frequency $p(w) \propto \frac{1}{f_w^a}$, where $a$ is a parameter
of the distribution which varies for different corpii. We estimate $a$ using maximum likelihood
estimation using the method in \cite{journals/siamrev/ClausetSN09} and compute the root-mean-squared error (RMSE)
between the estimated probability of a word according to the estimated Zipf distribution
and the empirical probability that word measured by counting word frequencies. A corpus
that resembles natural language more closely will have a low RMSE to its correpsonding
Zipf distribution. We find that the paraphrased datasets better fit their Zipf
distribution. We also visualize in both Figure \ref{fig:gscan_zipf} the ordered
frequency distribution of the paraphrased gSCAN dataset and its corresponding Zip
probability density function.

\subsection{Compositional Properties} 

We also examine whether the datasets maintained their compositional properties. Recall
that the datasets are stratified into different splits to test different compositional
generalization cases. We want to test whether these cases still hold. Clearly, in the
output space, the compositional stratification still holds because we do not change the
output actions. In the input space, we can only measure whether the same object
is mentioned in each synthetic instruction and its corresponding paraphrased instruction,
because the verbs and adverbs may be changed to a synonym or a sequence of words having
a similar meaning. 

\begin{table}[H]
\centering
\begin{tabular}{lrrr}
\toprule
{} &   Size &  Color &  Object \\
\midrule
gSCAN   & 100\% & 99.98\% &  98.63\% \\
SR      & 100\% & 100\% &  100\% \\
ReaSCAN & 100\% & 99.99\% &  99.93\% \\
\bottomrule
\end{tabular}
\caption{Percentage of examples in each training set whether the object mentioned in the synthetic dataset was also found in exactly the same way the corresponding paraphrased example.}
\label{tab:compositional_properties}
\end{table}

As shown in Table \ref{tab:compositional_properties}, the retainment of target objects
is very high, never going under 98\%. We can be confident that the correct target object
is mentioned in the same way in the paraphrased examples.

\section{Evaluation of baselines on Natural-ish gSCAN, gSCAN-SR and ReaSCAN}
\label{sec:appendix_baselines_natural_ish_gscan}
We evaluate current published state-of-the-art models with openly available code
on the new datasets using our own re-implementation. We calculate the exact-match
performance using seeds 0-9 using the same hyperparameters for each model, the
details of which are specified in Appendix \ref{sec:additional_comparisons}. The models
are briefly described below:

\paragraph{ViLBERT with Cross-Attention} The ViLBERT model proposed in \cite{conf/emnlp/QiuH0SS21}, with only cross-attention between visual and text input streams, then decoding
the target action sequence autoregressively. As in \cite{conf/emnlp/Sikarwar22}, the
multi-level CNN on the grid world is replaced by adding learnable position encodings.

\paragraph{VilBERT with GRoCoT Self-Attention} The same ViLBERT model but with the tweaks proposed in \cite{conf/emnlp/Sikarwar22}, namely self-attention layers before cross-attention
layers..

\paragraph{Encoder-Decoder Transformer} A standard encoder-decoder Transformer, where the
transformer input sequence is the position-encoded and embedded visual stream concatenated
with the instruction, and the target output sequence are the actions, decoded autoregressively.

\subsection{Results}

\begin{table}[H]
\centering
\begin{tabular}{lll|l}
\toprule
{} &      \footnotesize{Transformer} &      \footnotesize{ViLBERT} &  \footnotesize{ViLBERT(PP)} \\
\hline
\multicolumn{4}{c}{\footnotesize{gSCAN}} \\
\hline
A     &    1.0 \footnotesize{± 0.0} &    1.0 \footnotesize{± 0.0} &    1.0 \footnotesize{± 0.0} \\
B     &  0.86 \footnotesize{± 0.28} &  0.94 \footnotesize{± 0.11} &  0.93 \footnotesize{± 0.09} \\
C     &  0.89 \footnotesize{± 0.16} &  0.89 \footnotesize{± 0.13} &  0.82 \footnotesize{± 0.26} \\
D     &  0.01 \footnotesize{± 0.02} &   0.0 \footnotesize{± 0.01} &    0.0 \footnotesize{± 0.0} \\
E     &  0.99 \footnotesize{± 0.02} &  0.93 \footnotesize{± 0.12} &  0.71 \footnotesize{± 0.24} \\
F     &    1.0 \footnotesize{± 0.0} &    1.0 \footnotesize{± 0.0} &    1.0 \footnotesize{± 0.0} \\
G     &    0.0 \footnotesize{± 0.0} &    0.0 \footnotesize{± 0.0} &    0.0 \footnotesize{± 0.0} \\
H     &  0.19 \footnotesize{± 0.06} &  0.23 \footnotesize{± 0.01} &  0.17 \footnotesize{± 0.06} \\
\hline
\multicolumn{4}{c}{\footnotesize{gSCAN-SR}} \\
\hline
I  &    1.0 \footnotesize{± 0.0} &    1.0 \footnotesize{± 0.0} &    1.0 \footnotesize{± 0.0} \\
II    &  0.95 \footnotesize{± 0.04} &  0.93 \footnotesize{± 0.04} &  0.96 \footnotesize{± 0.02} \\
III   &  0.99 \footnotesize{± 0.01} &  0.96 \footnotesize{± 0.03} &    1.0 \footnotesize{± 0.0} \\
IV    &    1.0 \footnotesize{± 0.0} &    1.0 \footnotesize{± 0.0} &    1.0 \footnotesize{± 0.0} \\
V     &  0.46 \footnotesize{± 0.26} &   0.72 \footnotesize{± 0.1} &   0.9 \footnotesize{± 0.04} \\
VI    &  0.17 \footnotesize{± 0.18} &  0.61 \footnotesize{± 0.23} &  0.89 \footnotesize{± 0.06} \\
\hline
\multicolumn{4}{c}{\footnotesize{ReaSCAN}} \\
\hline
IID  &   0.99 \footnotesize{± 0.0} &  0.98 \footnotesize{± 0.02} &  0.97 \footnotesize{± 0.01} \\
A1    &  0.94 \footnotesize{± 0.02} &  0.95 \footnotesize{± 0.04} &  0.95 \footnotesize{± 0.01} \\
A2    &  0.61 \footnotesize{± 0.05} &  0.52 \footnotesize{± 0.13} &  0.46 \footnotesize{± 0.07} \\
B1    &  0.75 \footnotesize{± 0.02} &  0.79 \footnotesize{± 0.05} &  0.75 \footnotesize{± 0.03} \\
B2    &  0.54 \footnotesize{± 0.02} &   0.6 \footnotesize{± 0.09} &  0.53 \footnotesize{± 0.05} \\
C1    &  0.37 \footnotesize{± 0.02} &  0.32 \footnotesize{± 0.02} &  0.64 \footnotesize{± 0.03} \\
C2    &  0.27 \footnotesize{± 0.05} &  0.22 \footnotesize{± 0.05} &  0.22 \footnotesize{± 0.03} \\
\bottomrule
\end{tabular}
\caption{The evaluation results for gSCAN, gSCAN-SR and ReaSCAN at 300,000 iterations, where performance for
splits B-H is measured at the point where the model performed best on split A during training.
ViLBERT is the model in \cite{conf/emnlp/QiuH0SS21} and Tformer is an
Encoder-Decoder Transformer. Tformer(PP) the same Transformer architecture evaluated
on the paraphrased dataset.}
\label{tab:pp_baseline_results}
\end{table}

\section{Image-Based gSCAN}
\label{sec:appendix_gscan_img}

\begin{table}[t]
\setlength{\tabcolsep}{5pt}
% Note: Not yet complete
\begin{tabular}{c|ll|ll}
\hline
{} & \multicolumn{2}{c}{Transformer} & \multicolumn{2}{c}{DemoGen} \\
{} &           \multicolumn{1}{c}{NL} &       \multicolumn{1}{c}{+Img} &           \multicolumn{1}{c}{NL} &       \multicolumn{1}{c}{+Img} \\
\hline
A & \footnotesize{1.0} \footnotesize{± 0.0} & \footnotesize{1.0} \footnotesize{± 0.0} & \footnotesize{0.99} \footnotesize{± 0.0} & \footnotesize{0.84} \footnotesize{± 0.01} \\
B & \footnotesize{0.99} \footnotesize{± 0.0} & \footnotesize{0.93} \footnotesize{± 0.08} & \footnotesize{0.96} \footnotesize{± 0.0} & \footnotesize{0.53} \footnotesize{± 0.01} \\
C & \footnotesize{0.99} \footnotesize{± 0.03} & \footnotesize{0.89} \footnotesize{± 0.16} & \footnotesize{0.97} \footnotesize{± 0.0} & \footnotesize{0.54} \footnotesize{± 0.01} \\
D & \footnotesize{0.08} \footnotesize{± 0.16} & \footnotesize{0.0} \footnotesize{± 0.0} & \footnotesize{0.01} \footnotesize{± 0.01} & \footnotesize{0.11} \footnotesize{± 0.02} \\
E & \footnotesize{0.98} \footnotesize{± 0.03} & \footnotesize{0.83} \footnotesize{± 0.22} & \footnotesize{0.98} \footnotesize{± 0.0} & \footnotesize{0.67} \footnotesize{± 0.0} \\
F & \footnotesize{1.0} \footnotesize{± 0.0} & \footnotesize{1.0} \footnotesize{± 0.0} & \footnotesize{0.98} \footnotesize{± 0.0} & \footnotesize{0.88} \footnotesize{± 0.01} \\
G & \footnotesize{0.0} \footnotesize{± 0.0} & \footnotesize{0.0} \footnotesize{± 0.0} & \footnotesize{0.0} \footnotesize{± 0.0} & \footnotesize{0.0} \footnotesize{± 0.0} \\
H & \footnotesize{0.19} \footnotesize{± 0.03} & \footnotesize{0.06} \footnotesize{± 0.05} & \footnotesize{0.59} \footnotesize{± 0.06} & \footnotesize{0.48} \footnotesize{± 0.02} \\
\hline
\end{tabular}
\caption{Evaluation on natural language data. NL refers to natural language instructions, NL + Img refers to natural language instructions and patch-encoded images}
\label{tab:nl_img_results}
\end{table}

 We observed a similar boost on Split H for the NL + Img dataset as well. However, we note that the model for NL + Img
appeared to be underfitting, so it is possible that with
a larger model that the results could have been even better.

\section{Examples of generated demonstrations}
\label{sec:generated_demonstrations}

\begin{figure*}[ht]
\begin{subfigure}{\textwidth}
\resizebox{\textwidth}{!}{
    \begin{tikzpicture}[
        title/.style={font=\fontsize{6}{6}\color{black!50}\ttfamily},
        node distance = 10mm, 
    ]
    \node [fill=black!10,rounded corners, inner sep=3pt] (query) {
        \begin{tikzpicture}[node distance=0mm]
            \node [anchor=west] (title) {\footnotesize{Query}};
            \node [anchor=north, below=of title.south] (state)
        {\includegraphics[width=.35\textwidth,keepaspectratio=1,trim={4cm 1cm, 3.5cm 1cm},clip]{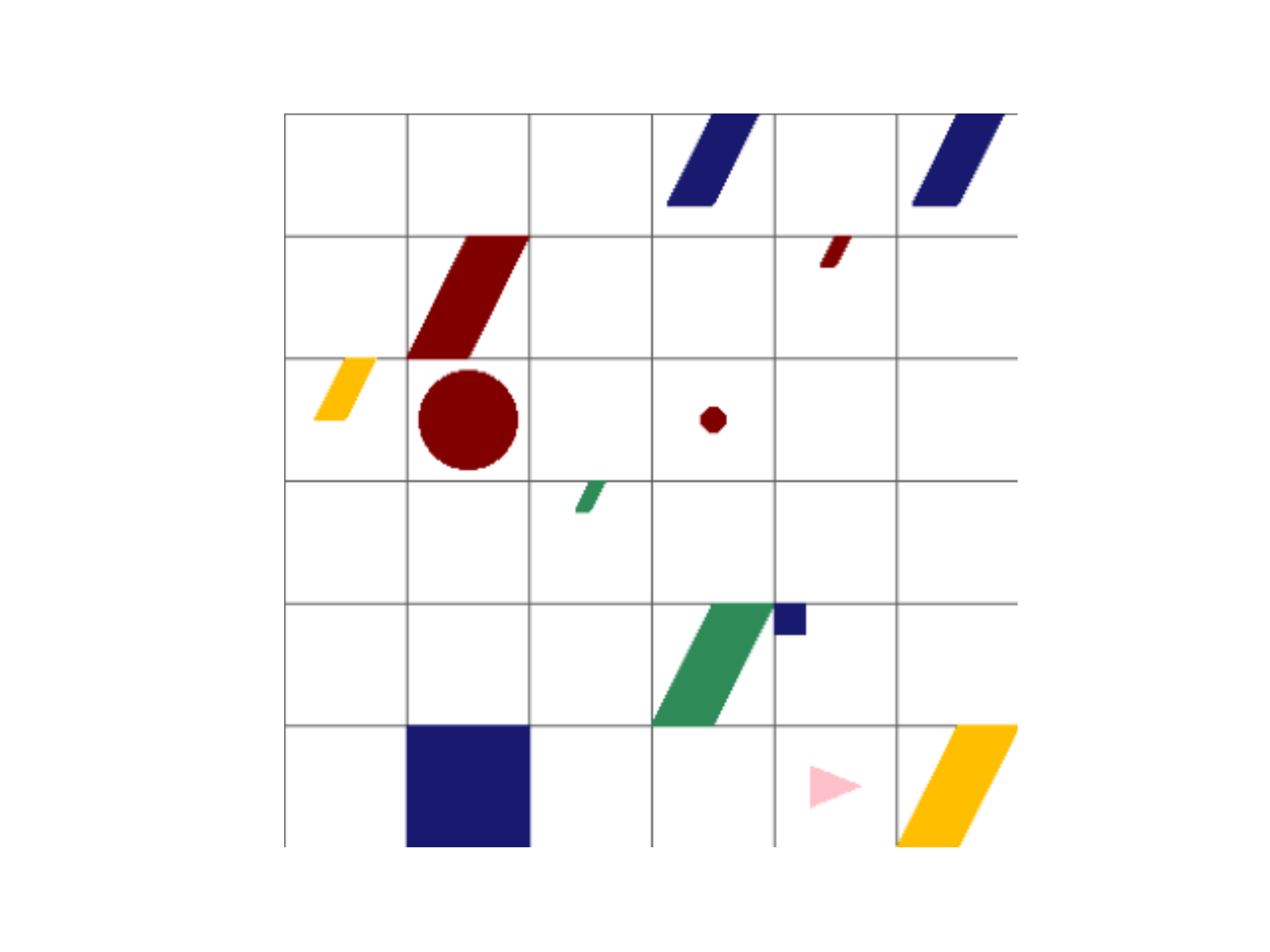}};
            \node[inner sep=0pt, below=of state.south] (query)
        {\footnotesize{$I_q$ = ``pull a red small circle while spinning"}};
        \end{tikzpicture}
    };
    \node [fill=black!10, rounded corners, inner sep = 5pt, right=of query.east, minimum width=50mm, anchor=south, rotate=90] (ig) {Instruction Generator};
    \node [fill=green!10, rounded corners, inner sep = 5pt, right= 7mm of ig.east, anchor=west, yshift=7mm] (i1) {\footnotesize{$I_1$ = ``pull a red small circle hesitantly"}};\node [fill=green!10, rounded corners, inner sep = 5pt, below=7mm of i1.west, anchor=west] (i2) {\footnotesize{$I_{2}$ = ``push a red big circle while spinning"}};
\node [fill=green!10, rounded corners, inner sep = 5pt, below=7mm of i2.west, anchor=west] (i3) {\footnotesize{$I_{3}$ = ``walk to a small circle hesitantly"}};
\node [fill=green!10, rounded corners, inner sep = 5pt, below=7mm of i3.west, anchor=west] (i4) {\footnotesize{$I_{4}$ = ``pull a circle hesitantly"}};
\node [fill=green!10, rounded corners, inner sep = 5pt, below=7mm of i4.west, anchor=west] (i5) {\footnotesize{$I_{5}$ = ``walk to a red circle hesitantly"}};
\node [fill=green!10, rounded corners, inner sep = 5pt, below=7mm of i5.west, anchor=west] (i6) {\footnotesize{$I_{6}$ = ``push a red big circle hesitantly"}};
\node [fill=green!10, rounded corners, inner sep = 5pt, below=7mm of i6.west, anchor=west] (i7) {\footnotesize{$I_{7}$ = ``pull a circle hesitantly"}};
\node [fill=yellow!10, rounded corners, inner sep = 5pt, below=7mm of i7.west, anchor=west] (i8) {\footnotesize{$I_{8}$ = ``pull a red small cylinder hesitantly"}};
\node [fill=yellow!10, rounded corners, inner sep = 5pt, below=7mm of i8.west, anchor=west] (i9) {\footnotesize{$I_{9}$ = ``walk to a small circle while spinning"}};
    \node [fill=black!10, rounded corners, inner sep = 5pt, right=9.5cm of query.east, minimum width=50mm, anchor=south, rotate=90] (at) {Transformer};
    \node [fill=green!10, rounded corners, inner sep = 5pt, right= 7mm of at.east, anchor=west, yshift=7mm] (a1) {\footnotesize{$A_1$ = ``LTURN(2) (WALK STAY)(3) RTURN (WALK STAY)(4)"}};
\node [fill=green!10, rounded corners, inner sep=5pt, below=7mm of a1.west, anchor=west] (a2) {\footnotesize{$A_{2}$ = ``LTURN(6) WALK LTURN(4) RTURN WALK (LTURN(4) WALK)(4)}"};
\node [fill=green!10, rounded corners, inner sep=5pt, below=7mm of a2.west, anchor=west] (a3) {\footnotesize{$A_{3}$ = ``LTURN(2) WALK STAY RTURN (WALK STAY)(3)"}};
\node [fill=green!10, rounded corners, inner sep=5pt, below=7mm of a3.west, anchor=west] (a4) {\footnotesize{$A_{4}$ = ``LTURN(2) WALK STAY RTURN (WALK STAY)(3) (PULL STAY)(3)"}};
\node [fill=green!10, rounded corners, inner sep=5pt, below=7mm of a4.west, anchor=west] (a5) {\footnotesize{$A_{5}$ = ``LTURN(2) (WALK STAY)(3) RTURN (WALK STAY)(3)"}};
\node [fill=green!10, rounded corners, inner sep=5pt, below=7mm of a5.west, anchor=west] (a6) {\footnotesize{$A_{6}$ = ``LTURN(2) (WALK STAY)(3) RTURN (WALK STAY)(3) (PUSH STAY)(4)"}};
\node [fill=green!10, rounded corners, inner sep=5pt, below=7mm of a6.west, anchor=west] (a7) {\footnotesize{$A_{7}$ = ``LTURN(2) (WALK STAY)(3) RTURN (WALK STAY)(3) (PULL STAY)(6)"}};
\node [fill=yellow!10, rounded corners, inner sep=5pt, below=7mm of a7.west, anchor=west] (a8) {\footnotesize{$A_{8}$ = ``LTURN(2) (WALK STAY)(4) RTURN (WALK STAY)(4)"}};
\node [fill=yellow!10, rounded corners, inner sep=5pt, below=7mm of a8.west, anchor=west] (a9) {\footnotesize{$A_{9}$ = ``LTURN(6) (WALK LTURN(4))(3) RTURN WALK (LTURN(4) WALK)(4)"}};
    \end{tikzpicture}
}
\caption{Support set generated by \textbf{Coverage Retrieval}}
\end{subfigure}

\begin{subfigure}{\textwidth}
\resizebox{\textwidth}{!}{
    \begin{tikzpicture}[
        title/.style={font=\fontsize{6}{6}\color{black!50}\ttfamily},
        node distance = 10mm, 
    ]
    \node [fill=black!10,rounded corners, inner sep=3pt] (query) {
        \begin{tikzpicture}[node distance=0mm]
            \node [anchor=west] (title) {\footnotesize{Query}};
            \node [anchor=north, below=of title.south] (state)
        {\includegraphics[width=.35\textwidth,keepaspectratio=1,trim={4cm 1cm, 3.5cm 1cm},clip]{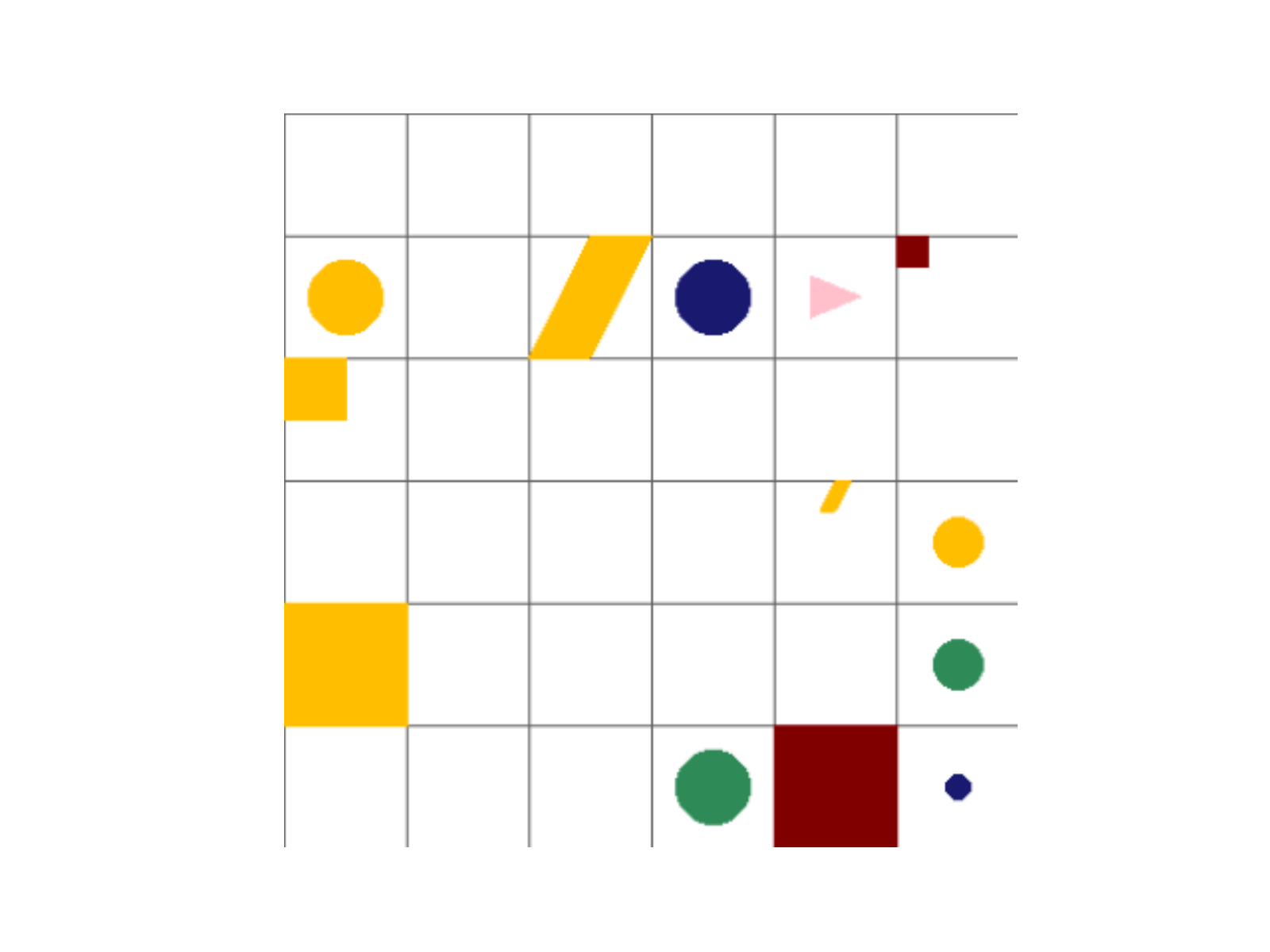}};
            \node[inner sep=0pt, below=of state.south] (query)
        {\footnotesize{$I^Q$ = ``pull a yellow cylinder while spinning"}};
        \end{tikzpicture}
    };
    \node [fill=black!10, rounded corners, inner sep = 5pt, right=of query.east, minimum width=50mm, anchor=south, rotate=90] (ig) {Instruction Generator};
    \node [fill=green!10, rounded corners, inner sep = 5pt, right= 7mm of ig.east, anchor=west, yshift=7mm] (i1) {\footnotesize{$I_1$ = ``pull a small cylinder"}};
\node [fill=green!10, rounded corners, inner sep = 5pt, below=7mm of i1.west, anchor=west] (i2) {\footnotesize{$I_{4}$ = ``pull a yellow small cylinder while zigzagging" }};
\node [fill=yellow!10, rounded corners, inner sep = 5pt, below=7mm of i2.west, anchor=west] (i3) {\footnotesize{$I_{14}$ = ``pull a small circle"}};
\node [fill=yellow!10, rounded corners, inner sep = 5pt, below=7mm of i3.west, anchor=west] (i4) {\footnotesize{$I_{15}$ = ``pull a big cylinder"}};
\node [fill=yellow!10, rounded corners, inner sep = 5pt, below=7mm of i4.west, anchor=west] (i5) {\footnotesize{$I_{16}$ = ``pull a big cylinder"}};
    \node [fill=black!10, rounded corners, inner sep = 5pt, right=9.5cm of query.east, minimum width=50mm, anchor=south, rotate=90] (at) {Transformer};
    \node [fill=green!10, rounded corners, inner sep = 5pt, right= 7mm of at.east, anchor=west, yshift=7mm] (a1) {\footnotesize{$A_1$ = ``LTURN(2) WALK PULL"}};
\node [fill=green!10, rounded corners, inner sep=5pt, below=7mm of a1.west, anchor=west] (a2) {\footnotesize{$A_{4}$ = ``LTURN(2) WALK RTURN WALK LTURN WALK PULL(2)}};
\node [fill=yellow!10, rounded corners, inner sep=5pt, below=7mm of a2.west, anchor=west] (a3) {\footnotesize{$A_{14}$ = ``LTURN(2) WALK PULL}};
\node [fill=yellow!10, rounded corners, inner sep=5pt, below=7mm of a3.west, anchor=west] (a4) {\footnotesize{$A_{15}$ = ``LTURN(2) WALK PULL}};
\node [fill=yellow!10, rounded corners, inner sep=5pt, below=7mm of a4.west, anchor=west] (a5) {\footnotesize{$A_{16}$ = ``LTURN(2) WALK PULL}};
    \end{tikzpicture}
    }
\caption{Support set generated by \textbf{GandR}}
\end{subfigure}

\begin{subfigure}{\textwidth}
\resizebox{\textwidth}{!}{
    \begin{tikzpicture}[
        title/.style={font=\fontsize{6}{6}\color{black!50}\ttfamily},
        node distance = 10mm, 
    ]
    \node [fill=black!10,rounded corners, inner sep=3pt] (query) {
        \begin{tikzpicture}[node distance=0mm]
            \node [anchor=west] (title) {\footnotesize{Query}};
            \node [anchor=north, below=of title.south] (state)
        {\includegraphics[width=.35\textwidth,keepaspectratio=1,trim={4cm 1cm, 3.5cm 1cm},clip]{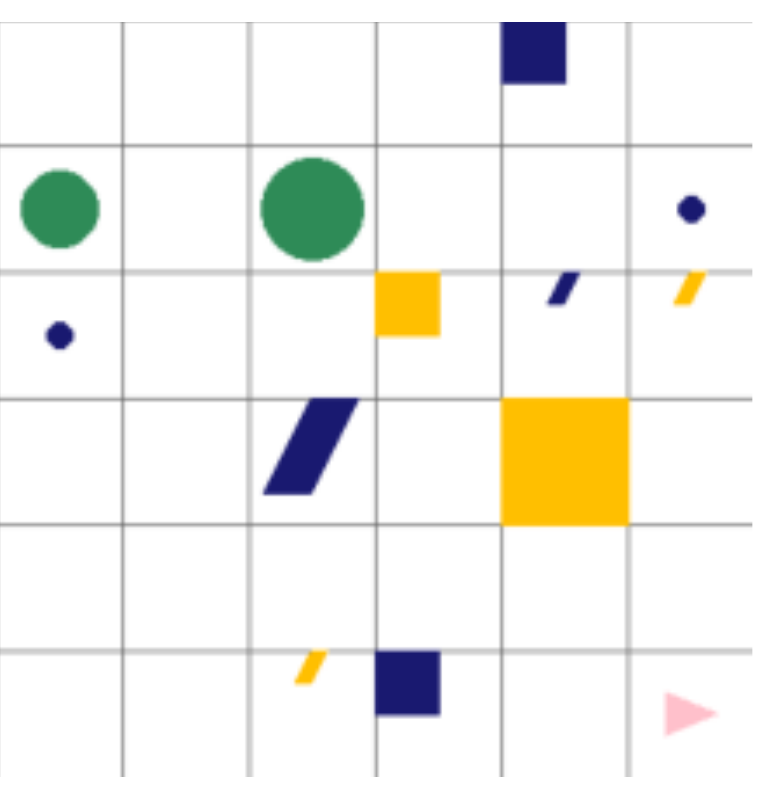}};
            \node[inner sep=0pt, below=of state.south] (query)
        {\footnotesize{$I_q$ = ``pull a green small circle while spinning"}};
        \end{tikzpicture}
    };
    \node [fill=black!10, rounded corners, inner sep = 5pt, right=of query.east, minimum width=50mm, anchor=south, rotate=90] (ig) {Instruction Generator};
    \node [fill=green!10, rounded corners, inner sep = 5pt, right= 7mm of ig.east, anchor=west, yshift=7mm] (i1) {\footnotesize{$I_1$ = ``walk to a green small circle while spinning"}};
\node [fill=green!10, rounded corners, inner sep = 5pt, below=7mm of i1.west, anchor=west] (i2) {\footnotesize{$I_2$ = ``push a green small circle while spinning}};
\node [fill=green!10, rounded corners, inner sep = 5pt, below=7mm of i2.west, anchor=west] (i3) {\footnotesize{$I_3$ = ``pull a green small circle while zigzagging}};
\node [fill=green!10, rounded corners, inner sep = 5pt, below=7mm of i3.west, anchor=west] (i4) {\footnotesize{$I_4$ = ``pull a green small circle hesitantly}};
\node [fill=green!10, rounded corners, inner sep = 5pt, below=7mm of i4.west, anchor=west] (i5) {\footnotesize{$I_5$ = ``pull a green small circle}};
    \node [fill=black!10, rounded corners, inner sep = 5pt, right=9.5cm of query.east, minimum width=50mm, anchor=south, rotate=90] (at) {Transformer};
    \node [fill=green!10, rounded corners, inner sep = 5pt, right= 7mm of at.east, anchor=west, yshift=7mm] (a1) {\footnotesize{$A_1$ = ``LTURN(6) (WALK LTURN(4))(5) RTURN (WALK LTURN(4))(3) WALK"}};
\node [fill=green!10, rounded corners, inner sep=5pt, below=7mm of a1.west, anchor=west] (a2) {\footnotesize{$A_2$ = ``LTURN(6) (WALK LTURN(4))(5) RTURN (WALK LTURN(4))(3) PUSH LTURN(4) PUSH}};
\node [fill=green!10, rounded corners, inner sep=5pt, below=7mm of a2.west, anchor=west] (a3) {\footnotesize{$A_3$ = ``LTURN(2) (WALK RTURN WALK LTURN)(4) WALK PULL(2)}};
\node [fill=green!10, rounded corners, inner sep=5pt, below=7mm of a3.west, anchor=west] (a4) {\footnotesize{$A_4$ = ``LTURN(2) (WALK STAY)(5) RTURN (WALK STAY)(4)}};
\node [fill=green!10, rounded corners, inner sep=5pt, below=7mm of a4.west, anchor=west] (a5) {\footnotesize{$A_5$ = ``LTURN(2) WALK(5) RTURN WALK(4)}};
    \end{tikzpicture}
    }
\caption{Support set generated by \textbf{Heuristic}}
\end{subfigure}

\begin{subfigure}{\textwidth}
\resizebox{\textwidth}{!}{
    \begin{tikzpicture}[
        title/.style={font=\fontsize{6}{6}\color{black!50}\ttfamily},
        node distance = 10mm, 
    ]
    \node [fill=black!10,rounded corners, inner sep=3pt] (query) {
        \begin{tikzpicture}[node distance=0mm]
            \node [anchor=west] (title) {\footnotesize{Query}};
            \node [anchor=north, below=of title.south] (state)
        {\includegraphics[width=.35\textwidth,keepaspectratio=1,trim={4cm 1cm, 3.5cm 1cm},clip]{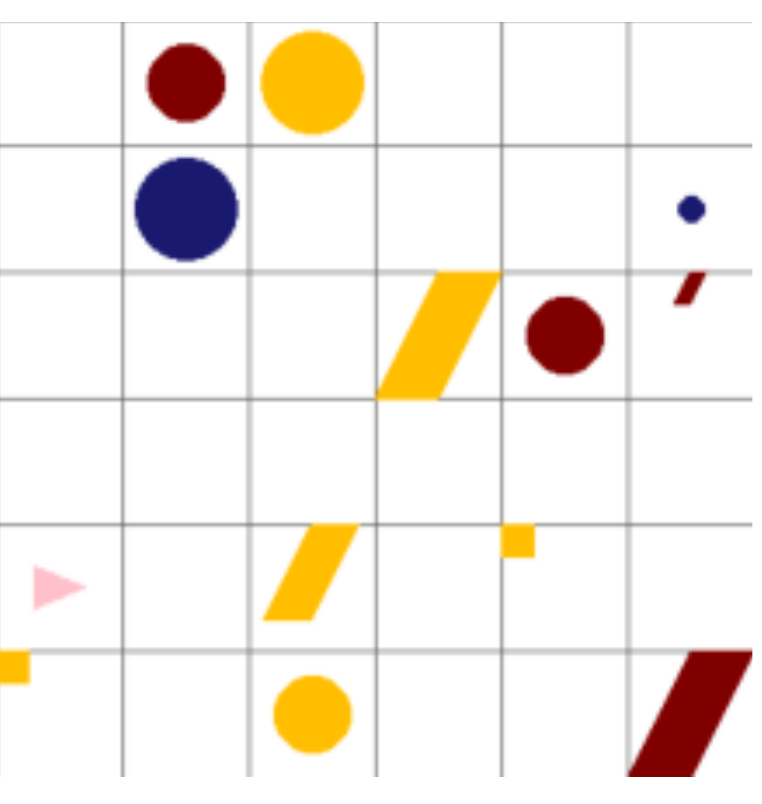}};
            \node[inner sep=0pt, below=of state.south] (query)
        {\footnotesize{$I_q$ = ``pull a blue small circle while spinning"}};
        \end{tikzpicture}
    };
    \node [fill=black!10, rounded corners, inner sep = 5pt, right=of query.east, minimum width=50mm, anchor=south, rotate=90] (ig) {Instruction Generator};
    \node [fill=green!10, rounded corners, inner sep = 5pt, right= 7mm of ig.east, anchor=west, yshift=7mm] (i1) {\footnotesize{$I_1$ = ``walk to a blue small circle while spinning"}};
\node [fill=green!10, rounded corners, inner sep = 5pt, below=7mm of i1.west, anchor=west] (i2) {\footnotesize{$I_2$ = ``push a blue small circle while spinning}};
\node [fill=green!10, rounded corners, inner sep = 5pt, below=7mm of i2.west, anchor=west] (i3) {\footnotesize{$I_3$ = ``pull a blue small circle while zigzagging}};
\node [fill=green!10, rounded corners, inner sep = 5pt, below=7mm of i3.west, anchor=west] (i4) {\footnotesize{$I_4$ = ``pull a blue small circle hesitantly}};
\node [fill=green!10, rounded corners, inner sep = 5pt, below=7mm of i4.west, anchor=west] (i5) {\footnotesize{$I_5$ = ``pull a blue small circle}};
    \node [fill=black!10, rounded corners, inner sep = 5pt, right=9.5cm of query.east, minimum width=50mm, anchor=south, rotate=90] (at) {Transformer};
    \node [fill=green!10, rounded corners, inner sep = 5pt, right= 7mm of at.east, anchor=west, yshift=7mm] (a1) {\footnotesize{$A_1$ = ``LTURN(4) (WALK LTURN(4))(4) RTURN (WALK LTURN(4))(3) WALK"}};
\node [fill=green!10, rounded corners, inner sep=5pt, below=7mm of a1.west, anchor=west] (a2) {\footnotesize{$A_2$ = ``LTURN(6) (WALK LTURN(4))(4) RTURN (WALK LTURN(4))(3) PUSH LTURN(4) PUSH}};
\node [fill=green!10, rounded corners, inner sep=5pt, below=7mm of a2.west, anchor=west] (a3) {\footnotesize{$A_3$ = ``LTURN WALK PULL(2)}};
\node [fill=green!10, rounded corners, inner sep=5pt, below=7mm of a3.west, anchor=west] (a4) {\footnotesize{$A_4$ = ``LTURN(2) (WALK STAY)(2) RTURN (WALK STAY)(4) (PULL STAY)(5)}};
\node [fill=green!10, rounded corners, inner sep=5pt, below=7mm of a4.west, anchor=west] (a5) {\footnotesize{$A_5$ = ``LTURN(2) WALK(4) RTURN WALK(4) PULL(10)}};
    \end{tikzpicture}
}
\caption{Support set generated by \textbf{Other States}}
\end{subfigure}

\begin{subfigure}{\textwidth}
\resizebox{\textwidth}{!}{
    \begin{tikzpicture}[
        title/.style={font=\fontsize{6}{6}\color{black!50}\ttfamily},
        node distance = 10mm, 
    ]
    \node [fill=black!10,rounded corners, inner sep=3pt] (query) {
        \begin{tikzpicture}[node distance=0mm]
            \node [anchor=west] (title) {\footnotesize{Query}};
            \node [anchor=north, below=of title.south] (state)
        {\includegraphics[width=.35\textwidth,keepaspectratio=1,trim={4cm 1cm, 3.5cm 1cm},clip]{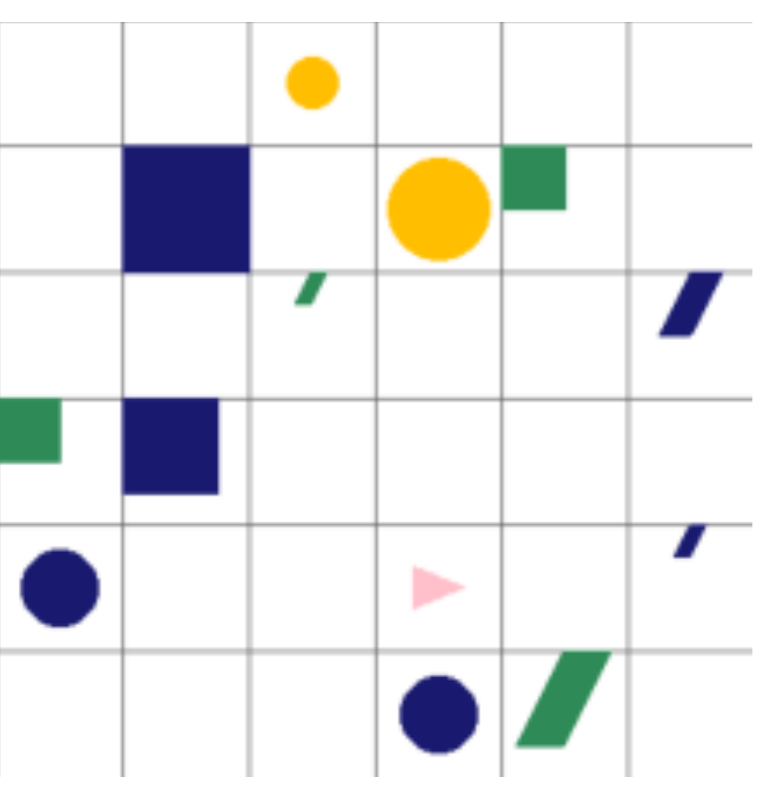}};
            \node[inner sep=0pt, below=of state.south] (query)
        {\footnotesize{$I_q$ = ``pull a blue small square while spinning"}};
        \end{tikzpicture}
    };
    \node [fill=black!10, rounded corners, inner sep = 5pt, right=of query.east, minimum width=50mm, anchor=south, rotate=90] (ig) {Instruction Generator};
    \node [fill=green!10, rounded corners, inner sep = 5pt, right= 7mm of ig.east, anchor=west, yshift=7mm] (i1) {\footnotesize{$I_1$ = ``push a big blue square while zigzagging"}};
\node [fill=green!10, rounded corners, inner sep = 5pt, below=7mm of i1.west, anchor=west] (i2) {\footnotesize{$I_2$ = ``push a big blue square while spinning}};
\node [fill=yellow!10, rounded corners, inner sep = 5pt, below=7mm of i2.west, anchor=west] (i3) {\footnotesize{$I_3$ = ``push a small yellow circle}};
\node [fill=yellow!10, rounded corners, inner sep = 5pt, below=7mm of i3.west, anchor=west] (i4) {\footnotesize{$I_4$ = ``push a big blue cylinder}};
\node [fill=yellow!10, rounded corners, inner sep = 5pt, below=7mm of i4.west, anchor=west] (i5) {\footnotesize{$I_5$ = ``walk to a small green cylinder while zigzagging}};
\node [fill=yellow!10, rounded corners, inner sep = 5pt, below=7mm of i5.west, anchor=west] (i6) {\footnotesize{$I_6$ = ``pull a big blue circle while spinning}};
\node [fill=yellow!10, rounded corners, inner sep = 5pt, below=7mm of i6.west, anchor=west] (i7) {\footnotesize{$I_7$ = ``push a big blue cylinder while spinning}};
\node [fill=yellow!10, rounded corners, inner sep = 5pt, below=7mm of i7.west, anchor=west] (i8) {\footnotesize{$I_8$ = ``pull a big blue cylinder}};
\node [fill=yellow!10, rounded corners, inner sep = 5pt, below=7mm of i8.west, anchor=west] (i9) {\footnotesize{$I_9$ = ``push a small yellow circle while zigzagging}};
    \node [fill=black!10, rounded corners, inner sep = 5pt, right=9.5cm of query.east, minimum width=50mm, anchor=south, rotate=90] (at) {Transformer};
    \node [fill=green!10, rounded corners, inner sep = 5pt, right= 7mm of at.east, anchor=west, yshift=7mm] (a1) {\footnotesize{$A_1$ = ``LTURN(2) WALK RTURN WALK LTURN WALK RTURN WALK(2) PUSH(2)"}};
\node [fill=green!10, rounded corners, inner sep=5pt, below=7mm of a1.west, anchor=west] (a2) {\footnotesize{$A_2$ = ``LTURN(6) (WALK LTURN(4))(2) RTURN (WALK LTURN(4))(3) PUSH LTURN(4) PUSH}};
\node [fill=yellow!10, rounded corners, inner sep=5pt, below=7mm of a2.west, anchor=west] (a3) {\footnotesize{$A_3$ = ``LTURN(2) WALK RTURN WALK(4)}};
\node [fill=yellow!10, rounded corners, inner sep=5pt, below=7mm of a3.west, anchor=west] (a4) {\footnotesize{$A_4$ = ``WALK(2) LTURN WALK(2) PUSH(2)}};
\node [fill=yellow!10, rounded corners, inner sep=5pt, below=7mm of a4.west, anchor=west] (a5) {\footnotesize{$A_5$ = ``LTURN(2) WALK RTURN WALK(2)}};
\node [fill=yellow!10, rounded corners, inner sep=5pt, below=7mm of a5.west, anchor=west] (a6) {\footnotesize{$A_6$ = ``LTURN(4) RTURN WALK (LTURN(4) PULL)(6) PULL}};
\node [fill=yellow!10, rounded corners, inner sep=5pt, below=7mm of a6.west, anchor=west] (a7) {\footnotesize{$A_7$ = ``(LTURN(4) WALK)(2) LTURN(5) (WALK LTURN(4))(2) PUSH LTURN(4) PUSH}};
\node [fill=yellow!10, rounded corners, inner sep=5pt, below=7mm of a7.west, anchor=west] (a8) {\footnotesize{$A_8$ = ``WALK(2) LTURN WALK(2) PULL}};
\node [fill=yellow!10, rounded corners, inner sep=5pt, below=7mm of a8.west, anchor=west] (a9) {\footnotesize{$A_9$ = ``LTURN(2) WALK RTURN WALK(4)}};
    \end{tikzpicture}
    }
\caption{Support set generated by \textbf{Random Instructions}}
\end{subfigure}
\caption{Demonstrations generated on Split H for different kinds of demonstration strategies.}
\label{fig:generation_examples}
\end{figure*}

We provide one-example-per-method of each support generation method on Split H in Figure \ref{fig:generation_examples}. Examples in green are valid in the environment, relevant to the target object and correctly executed. Examples in yellow are considered "not relevant" since they concern an object
with different properties than the one mentioned in the query. Examples in red are not correctly executed. Examples in grey are not valid in the environment. Note that for retrieval-based methods like \textbf{GandR} and \textbf{Retrieval}, the instruction is being solved in a different state to the
query one, which is the reason why the action trajectories are both valid and correct, but look very different from each other. Up to 9 of the 16 possible supports are shown.

Notice that \textbf{GandR} does not demonstrate the desired adverb ``while spinning" (\texttt{WALK(4)}, because it is only finding near neighbours of ``pull", which
happen only with \texttt{WALK} and \texttt{PUSH}.

\end{document}